%% file: accepted.tex
\pdfoutput=1
\documentclass{article}

\usepackage[final,main]{neurips_2026}

\usepackage[utf8]{inputenc}
\usepackage[T1]{fontenc}
\usepackage{hyperref}
\usepackage{etoolbox}
\usepackage{url}
\usepackage{booktabs}
\usepackage{amsfonts}
\usepackage{amsmath}
\usepackage{amssymb}
\usepackage{nicefrac}
\usepackage{microtype}
\usepackage{xcolor}
\usepackage{algorithm}
\usepackage{algorithmic}
\usepackage{graphicx}
\usepackage{subcaption}
\usepackage{wrapfig}
\usepackage{tabularx}
\usepackage{booktabs}
\usepackage{multirow}
\usepackage{makecell}
\usepackage{caption}
\usepackage{floatflt}
\usepackage{pifont}
\usepackage{tikz}
\usepackage{nicematrix}
\usepackage{floatflt}

\newcolumntype{Y}{>{\centering\arraybackslash}X}

\newcommand{\starwm}{\textsc{StarWM}}

\newcommand{\std}[1]{\tiny$\pm$#1}
\usepackage{nicematrix}

\title{\starwm: Self-Supervised Trained Attention Routing for Robust World Models}

\author{%
  Zeqiang Zhang$^{1}$, \ Fabian Wurzberger$^{1}$, \ Maximilian Otte$^{1}$, \ Daniel Schmid$^{1}$,\\
  \textbf{Sebastian Gottwald}$^{1}$, \ \textbf{Arne Peter Raulf}$^{2}$, \ \textbf{Daniel Alexander Braun}$^{1}$\\
  \rule{0pt}{14pt}$^{1}$Institute of Neural Information Processing, Ulm University \\ $^{2}$German Aerospace Center (DLR)\\
  \texttt{\{zeqiang.zhang, daniel.braun\}@uni-ulm.de}
}

\begin{document}
\setcitestyle{numbers, square}
\maketitle
\addtocontents{toc}{\protect\setcounter{tocdepth}{0}}
\begin{abstract}
  A robust world model must strike the balance between faithfully capturing environmental dynamics and abstracting away from irrelevant content. While reconstruction-based world models ensure faithful supervision, they misallocate representational capacity by pixel area rather than dynamics relevance for visual tasks, which can cause task-irrelevant content to dominate the learned representation. Alternatively, reconstruction-free methods avoid this bias but risk discarding possibly relevant information. We propose \starwm{}, which uses a cross-attention module trained on self-supervised dynamics to decide where reconstruction applies. A dual-stream decoder then restricts reconstruction to the attended regions, with stop-gradient barriers preventing interference between the two objectives. These components allows reconstruction to supervise the visual content of attended regions without contaminating the latent with non-predictive information. On DeepMind Control with dynamic video backgrounds, default (reward-free) \starwm{} achieves the strongest performance under random-frame distractors and substantially outperforms reconstruction-based baselines under sequential video. In addition, its reward-augmented variant matches or exceeds reconstruction-free methods on sequential video, achieving the highest overall return across all distractor regimes. Mechanistic probing confirms \starwm{} preserves state attributes with near-perfect fidelity through long-horizon imagination while systematically discarding distractors.  Code is available on GitHub\footnote{\url{https://github.com/zeqiang-zhang/StarWM}}.
\end{abstract}

\section{Introduction}
\label{sec:intro}

World models in the context of model-based reinforcement learning ideally allow simulating environmental dynamics in a compact latent space, enabling sample-efficient policy optimization~\citep{ha2018world}. Intuitively, a world model should capture environment dynamics with temporal coherence over long horizons, remain unaffected by task-irrelevant distractors, and align its representation with the overall objective, if available. 

Realizing these properties depends on how the representation is learned.
Reconstruction-based world models such as DreamerV3~\citep{hafner2025mastering} ground their representations through dense pixel reconstruction, but the gradient scales with pixel area rather than dynamical relevance, biasing the encoder toward whatever covers most of the frame (reconstruction bias)~\citep{deng2022dreamerpro,fu2021tia}. Reconstruction-free methods (DreamerPro~\citep{deng2022dreamerpro}, R2-Dreamer~\citep{morihira2026rdreamer}) replace reconstruction with auxiliary objectives such as latent consistency, contrastive prediction, or redundancy reduction. Self-supervised signals, such as inverse dynamics~\citep{pathak2017curiosity} and contrastive learning~\citep{laskin2020curl,schwarzer2021data}, provide meaningful gradients on action-conditional or temporally predictable regions, but capture only specific aspects of the underlying reality without pixel-level grounding~\citep{rakelly2021mutual}.

In the literature there have been multiple attempts to combine the pixel-level grounding of reconstruction with the flexibility of reconstruction-free methods that avoid the reconstruction bias in favor of a priori chosen auxiliary objectives.
For example, latent-level decomposition methods (TIA~\citep{fu2021tia}, DenoisedMDP~\citep{wang2022denoised}, Iso-Dream~\citep{pan2022iso}) allocate both reconstruction-based and -free signals to a shared latent representation. But this puts the two signals in constant gradient tension: reconstruction pushes the encoder to preserve all content, while the auxiliary objective pushes specific content into specific subspaces. The resulting allocation depends on carefully  balancing these conflicting gradients.

We propose \starwm{} (\textbf{S}elf-supervised \textbf{T}rained \textbf{A}ttention
\textbf{R}outing \textbf{W}orld \textbf{M}odel), which resolves the conflict between the two signals by architectural separation. \starwm{} performs the
decomposition on the encoder's spatial feature map and isolates the supervision sources
at the parameter level through attention routing. Our contributions are:

\begin{itemize}
    \item \textbf{A dual-stream architecture with self-supervised attention routing} that separates the where-to-attend decision from the what-to-extract decision. A cross-attention module produces a spatial mask, supervised exclusively by auxiliary objectives, with stop-gradient barriers ensuring that the two supervision sources update disjoint parameter subsets. The dual-stream decoder uses the spatial mask to route relevant content to the latent dynamics model, while the rest is passed to a residual stream that absorbs irrelevant content.

    \item \textbf{NoisyShape, a diagnostic environment} for mechanistic analysis of learned representations under visual distractions, providing controllable, easily interpreted dynamics and randomized video distractors. Ground-truth labels for color, shape, size, and position enable probe analysis of both immediate encoding and long-horizon imagination. 
            \item \textbf{Comprehensive qualitative and quantitative evaluations} of \starwm{}'s behavior under visual distractions. With the same hyperparameters, \starwm{} on average outperforms all baselines across distractor setups. On NoisyShape (reward-free), attention-based routing remains effective while alternatives degrade. Probe analysis confirms that \starwm{} retains task-relevant attributes while discarding irrelevant ones, achieving near-perfect temporal coherence over long horizons. Cross-sample image composition directly visualizes the three components' specialization without any explicit disentanglement objective.
\end{itemize}

\section{Preliminaries}
\label{sec:prelim}

\paragraph{Problem Setting.}

We consider a visual control task where the observation $o_t \in \mathbb{R}^{H \times W \times 3}$ is generated from two unobserved state components: a \textit{relevant} state that the world model should faithfully capture, and an \textit{irrelevant} state that should be filtered out. What counts as relevant depends on the chosen objective.  Given any such objective, our goal is a general world model whose latent dynamics capture the \textit{relevant} state while remaining uncontaminated by the \textit{irrelevant} state.

\paragraph{Reconstruction-Based World Models.}

Reconstruction-based world models learn a latent dynamics model by training an encoder--decoder architecture to reconstruct observations while predicting forward in latent space. A representative example is  DreamerV3~\citep{hafner2025mastering},  a world model centered on the Recurrent State Space Model (RSSM):
\begin{align}
    \text{Sequence model:} \quad & h_t = f_\theta(h_{t-1}, z_{t-1}, a_{t-1}), \\
    \text{Posterior:} \quad & z_t \sim q_\theta(z_t \mid h_t, o_t), \\
    \text{Prior:} \quad & \hat{z}_t \sim p_\theta(\hat{z}_t \mid h_t),
\end{align}
where $h_t$ is a deterministic recurrent state and $z_t$ a stochastic posterior inferred from the observation. The model is trained by maximizing an Evidence Lower Bound:

\begin{equation}
  \resizebox{0.94\linewidth}{!}{
  $
    \mathcal{L}_{\text{WM}} = \mathbb{E}\Big[
        \underbrace{-\ln p_\theta(o_t \mid h_t, z_t)}_{\text{Image loss }(\mathcal{L}_{\text{img}})}
        \underbrace{-\ln p_\theta(r_t \mid h_t, z_t)}_{\text{Reward loss }(\mathcal{L}_{\text{reward}})}
         \underbrace{-\ln p_\theta(c_t \mid h_t, z_t)}_{\text{Cont.\ loss }(\mathcal{L}_{\text{cont}})}
        + \underbrace{\beta \, \mathrm{KL}\!\big[\, q_\theta(\cdot \mid h_t, o_t) \,\|\, p_\theta(\cdot \mid h_t) \,\big]}_{\mathcal{L}_{\text{KL}}^{\text{RSSM}}} 
    \Big].
    $}
    \label{eq:elbo}
\end{equation}

To learn a policy, DreamerV3 trains an actor-critic agent on latent trajectories imagined by the RSSM. The actor maximizes the critic's estimated return over rollouts from encoded states. 

\section{Method: \starwm{}}
\label{sec:method}

\begin{figure}[t]
    \centering
    \includegraphics[width=\textwidth]{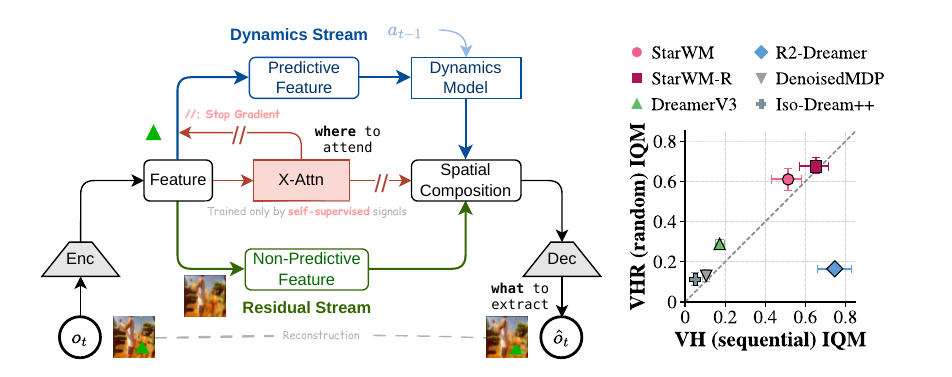}
    \caption{\textbf{\starwm{} disentangles \emph{where-to-attend} (X-Attn) from \emph{what-to-extract} (reconstruction $\hat{o}_t$).} \textbf{\textit{Left:}} Self-supervised objective signals train a cross-attention module to localize predictable, action-conditional content, while the content of the attended regions is shaped by a reconstruction objective through a dual-stream decoder. Stop-gradient barriers (//) keep the two objectives from interfering. For detailed architecture diagram see Appendix~\ref{app:algorithm}. \textbf{\textit{Right:}} Interquartile mean (IQM) aggregated across 6 DMC tasks under sequential (VH) and random (VHR) video distractors, normalized by DreamerV3 clean performance (recovery ratio). \starwm{} and its variant are the only methods robust to both distractor types.}
    \label{fig:starwm}
\end{figure}

\starwm{} addresses the reconstruction bias with a minimal modification 
to existing reconstruction-based world models: only the encoder and 
decoder are changed, while the latent dynamics model and policy 
learning procedure are inherited unchanged. The modification realizes two design principles, illustrated in Figure~\ref{fig:starwm}: \textbf{(1) Dual-stream routing.} A cross-attention module splits the encoder's spatial features into a dynamics stream (entering the latent dynamics model) and a residual stream (auxiliary reconstruction only, discarded during imagination), preventing the dynamics latent from encoding objective-irrelevant content. \textbf{(2) Self-supervised location/content separation.} Self-supervised signals train the attention to decide \emph{where} to route, while reconstruction trains the decoder to decide \emph{what} to extract. Stop-gradient barriers keep the two objectives from interfering.

Though we instantiate them on DreamerV3, these principles are independent of the specific dynamics backbone. Specifically, instead of applying the RSSM to the raw observations $o_t$, our approach applies it to the attention gated features $e_t$. The following subsections detail each component.

\subsection{Cross-Attention Spatial Routing}
\label{sec:where}
\starwm{} intercepts the encoder at an intermediate feature map of shape $C \times H \times W$ (see Appendix~\ref{app:algorithm}), flattened into $L = HW$ spatial tokens $F_t \in \mathbb{R}^{L \times C}$. As attention is permutation-invariant, we form keys by adding 2D sinusoidal positional encodings $E_\mathrm{pos} \in \mathbb{R}^{L \times C}$~\citep{carion2020detr} to enable location-aware attention, and values by appending normalized spatial coordinates  $\mathrm{E_\mathrm{coord}} \in \mathbb{R}^{L \times 2}$~\citep{locatello2020object,coord2018} to preserve explicit positional information for the dynamics model. Specifically, we set $K_t = F_t + E_\mathrm{pos}$ and $V_t = [F_t ;\; E_\mathrm{coord}]$.

To track distinct entities or articulated body parts independently, we employ $N$ learnable query vectors $Q \in \mathbb{R}^{N \times C}$. The queries attend over these features via scaled dot-product attention:
\begin{equation}
    A_t = \mathrm{Softmax}\!\left(\frac{(QW_Q)(K_tW_K)^\top}{\sqrt{C}}\right) \in \mathbb{R}^{N \times L},
    \label{eq:attn}
\end{equation}
where $W_Q, W_K \in \mathbb{R}^{C \times C}$ are learnable projections. Each row of $A_t$ is a soft spatial probability mask localizing one entity. We use a multi-head implementation in practice (Appendix~\ref{app:mha_detail}).

Crucially, $A_t$ routes the visual features into two complementary streams: a \emph{dynamics stream} of attended entities passed to the RSSM, and a \emph{residual stream} (handled by an auxiliary VAE) that aids reconstruction but is discarded during imagination. The following sections detail how this attention is trained (\S\ref{sec:attn_guidance}) and decoded (\S\ref{sec:what}).

\subsection{Self-Supervised Location/Content Separation}
\label{sec:attn_guidance}

We train the cross-attention module using self-supervised signals that each instantiate a notion of relevance grounded in environment dynamics. To compute these losses, we extract coordinate-free attention-pooled features $e_t^{\text{attn}} = \mathrm{flatten}(A_t F_t)$. Omitting spatial coordinates forces it to localize entities based purely on visual dynamics rather than absolute positions.

\paragraph{Inverse dynamics.}
A learnable inverse dynamics mapping~\citep{pathak2017curiosity} $\phi_\mathrm{inv}$ predicts the action $a_t$ from consecutive attention-pooled features, trained by optimizing
\begin{equation}
    \mathcal{L}_{\text{inv}} = \ell\!\big(\phi_\mathrm{inv}(e_t^{\text{attn}},\, e_{t+1}^{\text{attn}}),\; a_t\big),
    \label{eq:inv}
\end{equation}
where $\ell$ denotes cross-entropy or squared error depending on the action space. This objective guides attention to regions whose visual changes are caused by the agent's actions, while action-independent content produces no useful gradient.

\paragraph{Temporal contrastive learning.} A learnable projection mapping $\phi_\mathrm{ctr}$ is trained by
temporal contrastive learning~\citep{oord2018infonce, laskin2020curl}, pulling consecutive features together and pushing apart random negatives, resulting in the loss
\begin{equation}
    \mathcal{L}_{\text{ctr}} = -\log\sigma\!\left(\tfrac{\phi_\mathrm{ctr}(e_t^{\text{attn}})^\top \phi_\mathrm{ctr}(e_{t+1}^{\text{attn}})}{\tau}\right) - \log\sigma\!\left(-\tfrac{\phi_\mathrm{ctr}(e_t^{\text{attn}})^\top \phi_\mathrm{ctr}(\tilde e^{\text{attn}})}{\tau}\right),
    \label{eq:ctr}
\end{equation}
where $\tau$ a temperature, and $\tilde e^{\text{attn}}$ a random negative. This guides attention toward temporally predictable content.

Together, these signals route attention toward controllable and predictable entities, leaving the rest as irrelevant. While they serve as our default, the cross-attention pathway is agnostic to the choice of signal; for instance, adding reward prediction yields \starwm{}-R, which routes by task-relevance and better tracks passively moving task-relevant entities (see Appendix~\ref{app:reward_extension}).

\paragraph{Gradient isolation.}
All downstream components access a stop-gradiented copy of the attention map \smash{$\bar{A}_t = \mathrm{sg}(A_t)$}, numerically identical to $A_t$ but carrying no gradient. The entity tokens \smash{$e_t^{\text{dyn}}\in \mathbb{R}^{N \times (C+2)}$} entering the RSSM are computed as \smash{$e_t^{\text{dyn}} = \bar{A}_t  V_t$}, as the model's estimate of \emph{relevant} state. Additionally, the dual-stream decoder (\S\ref{sec:what}) uses $\bar{A}_t$ as the spatial routing mask. The attention parameters $\{Q, W_Q, W_K\}$ are trained purely by the self-supervised signals above, while the encoder continues to receive gradients from all three sources: self-supervised losses through  $A_t$, RSSM losses through $V_t$, and reconstruction losses through the decoder.

\subsection{Dual-Stream Decoder}
\label{sec:what}
With the attention map fixed by the stop-gradient barrier, the decoder composes the two streams into a reconstruction target that allows gradients to shape only entity content without altering the routing.

\paragraph{Dynamics stream via spatial broadcast.}
Following the spatial broadcast decoder design~\citep{watters2019spatial}, we first decode the dynamics features into predicted representations $\hat{e}_t = \phi_{\text{fg}}([h_t;z_t])$. We then render each representation back to its attended spatial positions by inverse broadcasting, computing \smash{$\hat{F}_t^{\text{fg}} = \bar{A}_t^\top \hat{e}_t$} with $\bar{A}_t$ as the composition mask. This ``paints'' each attended region back onto its spatial positions. The decoder reads the content entirely from dynamic features, while the attention map specifies only where to draw. This forces the RSSM to retain rich attribute information.

\paragraph{Residual stream.}

Parallel to the dynamics stream, the full encoder output $F_t$ is passed through a residual mapping $\phi_\mathrm{bg}$ producing the background features \smash{$\hat{F}_t^{\text{bg}}$}. In our implementation $\phi_\mathrm{bg}$ uses a Gaussian bottleneck regularized by unit Gaussian prior minimizing \smash{$\beta_{\text{bg}}\mathcal{L}_{\text{KL}}^{\text{bg}}$} similar to a VAE. The stochastic bottleneck prevents the residual stream from encoding the entire scene. Instead, it absorbs objective-irrelevant background details to satisfy the dense reconstruction objective. The background latent is not fed into the RSSM and is discarded during imagination rollouts. The dynamics latent handles only attended content, while the reconstruction of objective-irrelevant content is absorbed by the residual stream.

\paragraph{Spatial composition.}
The aggregate coverage mask $M_t = \mathrm{clamp}(\sum_n \bar{A}_{t,n},\,0,\,1)$ measures total objective-relevant content presence at each spatial location by summing over the $N$ queries. Because the attention in Eq.~\ref{eq:attn} is normalized over spatial tokens rather than over queries, individual query maps are not mutually exclusive and can place attention on shared locations; clamping ensures $M_t$ remains a valid blending weight in $[0,1]$. The two streams are merged at the feature level:
\begin{equation}
    \hat{F}_t = \hat{F}_t^{\text{fg}} \odot M_t \;+\; \hat{F}_t^{\text{bg}} \odot (1-M_t).
    \label{eq:compose}
\end{equation}
A shared transposed convolution upsamples $\hat{F}_t$ back into observation space, resulting in the reconstruction loss $\mathcal L_\mathrm{img}$. Composing at the feature level rather than at pixel level prevents the soft mask from implicitly encoding shape through its spatial extent~\citep{locatello2020object}.

Combining all components, the overall training objective for \starwm{} is:
\begin{equation}
    \mathcal{L} = \underbrace{\mathcal{L}_{\text{img}} + \mathcal{L}_{\text{reward}} + \mathcal{L}_{\text{cont}} + \mathcal{L}_{\text{KL}}^{\text{RSSM}}}_{\text{Dreamer backbone}} + \underbrace{\lambda_{\text{inv}}\,\mathcal{L}_{\text{inv}} + \lambda_{\text{ctr}}\,\mathcal{L}_{\text{ctr}}}_{\text{X-Attention guidance}} + \underbrace{\beta_{\text{bg}}\,\mathcal{L}_{\text{KL}}^{\text{bg}}}_{\text{Residual regularization}}.
    \label{eq:loss}
\end{equation}

\section{Experiments}
\label{sec:experiments}

Our evaluation addresses four questions about \starwm{}'s design: whether the
latent representation cleanly separates objective-relevant from
objective-irrelevant content (\S\ref{sec:exp_diagnostic}); whether this
separation translates into policy robustness across distractor types
(\S\ref{sec:exp_dmcgb}); what should train the attention, and whether isolating it from reconstruction is necessary (\S\ref{sec:exp_ablation}); and under which conditions the
default self-supervised routing signal suffices, and when task reward is needed to complete it (\S\ref{sec:exp_boundary}).

\paragraph{Baselines and implementation setup.} 

 \begin{wrapfigure}{r}{0.5\textwidth}
 \vspace{-20pt}
    \centering

    \includegraphics[width=0.5\textwidth]{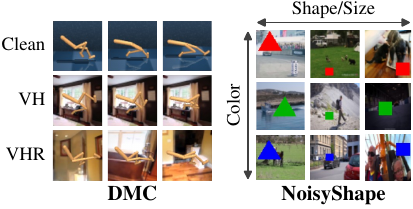}
    \caption{\textbf{Evaluation environments.} \textbf{\textit{Left:}} DeepMind Control Suite with three background conditions: clean, Video Hard Sequential, and Video Hard Random. \textbf{\textit{Right:}} NoisyShape: a geometric shape with controllable attributes (shape, color, size) on DAVIS video backgrounds, used for mechanistic analysis with ground-truth attribute probing; position is randomly re-sampled each step. Arrows denote the four possible actions.  }
    \label{fig:env_half}
\end{wrapfigure}

We compare \starwm{} against four baselines representing distinct distraction-handling paradigms: \textbf{DreamerV3}~\citep{hafner2025mastering}, our reconstruction-based backbone; \textbf{DenoisedMDP}~\citep{wang2022denoised}, latent decomposition based on controllability and reward-relevance; \textbf{Iso-Dream++}~\citep{pan2023isodreampp}, which splits the RSSM into controllable and non-controllable branches via inverse dynamics; and \textbf{R2-Dreamer}~\citep{morihira2026rdreamer}, a concurrent reconstruction-free approach replacing the pixel decoder with a redundancy-reduction objective.  Our evaluation uses two complementary environments: a controlled NoisyShape diagnostic environment with ground-truth attribute labels for mechanistic analysis, and the DeepMind Control Suite with video distractors for policy benchmarking. All methods use identical observation resolutions and training budgets (see Appendix~\ref{app:baseline}); all results report mean and standard deviation across 3 random seeds.

\subsection{Latent Separation: Probe Analysis on NoisyShape}
\label{sec:exp_diagnostic}

We propose NoisyShape as a diagnostic environment for mechanistic analysis of representation learning under controlled distractors (Figure~\ref{fig:env_half}; Appendix~\ref{app:shapeenv}). For each state, a geometric shape appears across randomly sampled DAVIS~\citep{perazzi2016benchmark} video backgrounds. The environment is constructed around three properties targeting distinct challenges: (i) \emph{objective-relevant attributes}, where shape and size are coupled while color cycles independently, all driven by action; (ii) \emph{objective-irrelevant noise}, including randomly re-sampled spatial positions and random video backgrounds that carry no predictive information; and (iii) a \emph{reward-free setup} that isolates representation learning from policy optimization. Combined with ground-truth labels for probing, these properties make NoisyShape suitable for two complementary analyses: evaluating baseline representation paradigms, and conducting incremental architectural ablations on the DreamerV3 backbone.

\begin{figure}[t]
    \centering
    \includegraphics[width=0.85\textwidth]{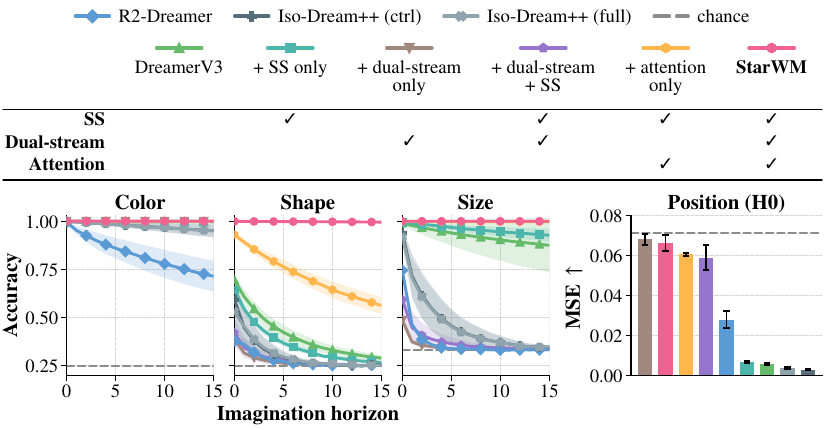}

    \caption{\textbf{Latent probe results on NoisyShape.} Probe accuracy for objective-relevant attributes (color, shape, size) from H0 (encoding) through H15 (15 steps of imagination), and objective-irrelevant attribute (position MSE) at H0 (H15 omitted as ground-truth position at $t{+}15$ is independent of the state at $t$; higher MSE indicates correct irrelevant treatment); dashed line = chance level.}
    \label{fig:probe_curves}
\end{figure}

We train probes on the RSSM latent to predict color, shape, and size at two stages: the encoding step (H0) and after 15 steps of open-loop imagination (H15). H0 measures whether the encoder captures objective-relevant attributes from observations; H15 tests whether the dynamics model preserves these attributes in latent space without further visual input. A representation contaminated by irrelevant content typically encodes well at H0 but loses the relevant signal during long-horizon rollouts, since the dynamics model cannot predict irrelevant noise. As NoisyShape is reward-free, our cross-method comparison here is at the level of latent-probe accuracy rather than policy return; policy comparisons are conducted on DMC (\S\ref{sec:exp_dmcgb}). Full evaluation protocol is in Appendix~\ref{app:probe_method} and evaluation on the variants of the NoisyShape is in Appendix~\ref{app:progressive_distraction}.

\paragraph{Objective-relevant attribute retention.}
Figure~\ref{fig:probe_curves} shows that \starwm{} achieves near-perfect retention across all relevant attributes through imagination, while every baseline degrades. DreamerV3's shape retention collapses through imagination, confirming the reconstruction bias; R2-Dreamer degrades already at encoding, since its redundancy-reduction objective lacks signal under random backgrounds; Iso-Dream++'s action-driven separation operates at the latent level and likewise fails to filter irrelevant content; and DenoisedMDP cannot be evaluated here as its decomposition strictly requires reward signals. To isolate which architectural component closes this gap, we evaluate ablations of \starwm{} that progressively add self-supervised attention guidance, the dual-stream decoder, and cross-attention routing to the DreamerV3 backbone. Every ablation configuration degrades on at least one attribute, indicating that all three components are necessary for distraction-robust representation learning in this reward-free setting.

\paragraph{Position as a diagnostic feature.}

The position attribute provides a test of selective encoding: an ideal dynamics latent encodes the attribute only when it is dynamically predictable. Though position is sampled uniformly at each step, DreamerV3 faithfully encodes position at H0 (near-zero MSE), driven by its dense reconstruction objective. \starwm{} discards most of this information (high MSE), correctly treating the random noise as irrelevant (Fig.~\ref{fig:probe_curves}, right). This filtering is driven by the dual-stream decoder: ablations without it retain position precisely. To rule out that \starwm{} simply drops spatial information regardless of its predictability, we evaluate on a variant where the object follows a predictable trajectory (Appendix~\ref{app:predictable_position}). In this variant, \starwm{} successfully retains object's position through 15 steps of imagination, confirming that the RSSM preserves spatial features when they are dynamically meaningful.

\paragraph{Cross-sample composition.}

\begin{figure}[t]
    \centering

    \includegraphics[width=0.85\textwidth]{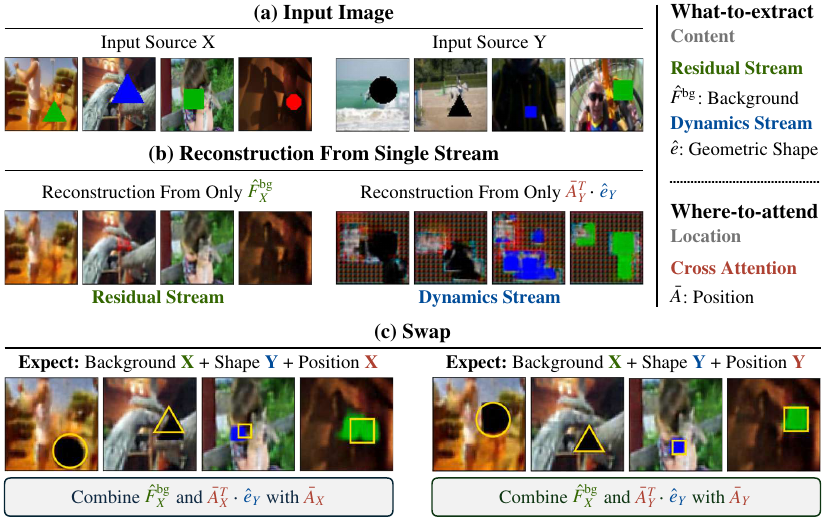}
    \caption{\textbf{Cross-sample composition reveals emergent factorization.} 
    \textbf{(a.)} Input frames from sources X and Y, with \textbf{(b.)} each stream's isolated reconstruction beneath. The residual stream renders only background context, while the dynamics stream renders entity content. 
\textbf{(c.)} Swapping components across X and Y produces controlled outputs: Y's entity on X's background, with position determined by whichever attention mask ($\bar{A}_X$ or $\bar{A}_Y$) is used. The three components specialize into distinct emergent roles, mapping onto the content and location separation of \S\ref{sec:method} without any explicit disentanglement objective.}

    \label{fig:composition}
\end{figure}

To directly show how the three architectural components specialize, we decode composite images that mix each component across two source frames (Figure~\ref{fig:composition}). Each substitution cleanly swaps the corresponding visual factor while leaving the others intact, revealing an emergent three-way factorization: the residual stream carries background context, the dynamics stream carries entity content, and the attention map carries spatial location. This factorization arises from the architectural design alone, without any explicit disentanglement objective.

\subsection{Policy Performance: DMC under Video Distractors}
\label{sec:exp_dmcgb}
\begin{figure}[t]
    \centering
    \includegraphics[width=\textwidth]{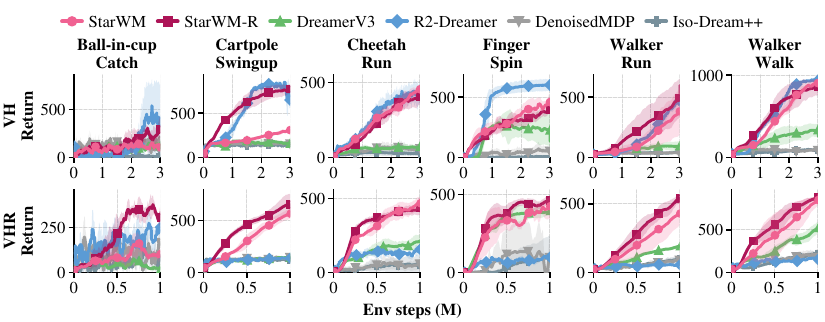}
    \caption{\textbf{DMC learning curves.} \starwm{} and \starwm{}-R are the only methods that perform strongly under both sequential (VH) and random (VHR) video backgrounds. 
    }
    \label{fig:dmc_curves}
\end{figure}

Figure~\ref{fig:dmc_curves} reports per-task evaluation returns on DMC under both video distractor settings, and Figure~\ref{fig:starwm} (right) summarizes cross-condition robustness via IQM aggregated across the six tasks, normalized to DreamerV3 clean performance, yielding a recovery ratio measuring how much of the clean-task return each method retains under distractors. \starwm{}'s additional architectural components do not come at a cost in clean environments, where it matches DreamerV3. Full numerical results, together with aggregate statistics under 95\% stratified bootstrap CIs, performance profiles, and pairwise probability-of-improvement heatmaps following~\citet{agarwal2021deep}, are in Appendix~\ref{app:full_results}. See Appendix~\ref{app:hparam_sensitivity} for hyperparameter sensitivity.

 \begin{wrapfigure}{r}{0.4\textwidth}
    \centering

    \includegraphics[width=0.4\textwidth]{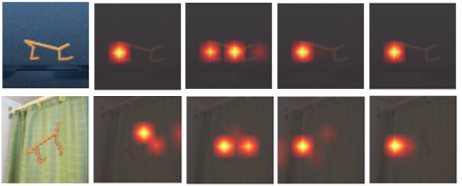}
    \caption{\textbf{Learned attention maps on DMC.} \starwm{}'s cross-attention consistently localizes the agent through 4 queries under both clean and distracting settings. }
    \label{fig:attention_dmc}
\end{wrapfigure}

Figure~\ref{fig:attention_dmc} visualizes \starwm{}'s attention maps 
under both clean and distractor (VH) settings: queries lock to the agent's body 
parts and do not drift to the visually dominant background. This is 
direct mechanistic evidence that the routing decision is governed by 
self-supervised signals rather than reconstruction's pixel. Appendix~\ref{app:separability} confirms that routing stays stable across matched textures, low-contrast agents, and photorealistic scenes. 

 Under random distractors (VHR), where temporal coherence is destroyed, default \starwm{} achieves the strongest performance on 5 of 6 tasks with substantial margins. This demonstrates the robustness of purely self-supervised cross-attention routing under unstructured noise, where baseline representations struggle to isolate the agent. Under sequential distractors (VH), R2-Dreamer leads on most tasks, since temporally coherent backgrounds align well with redundancy reduction. This same coherence partially co-opts \starwm{}'s contrastive guidance, which rewards any temporally consistent feature and cannot by itself disambiguate agent from background (Appendix~\ref{app:ss_ablation}); despite this, default \starwm{} remains competitive on high-DOF locomotion (Walker Walk, Cheetah Run), where inverse dynamics provides a strong agent-localizing signal.

In this regime where self-supervised signals underdetermine the routing target, the modular attention pathway becomes useful. As our core routing architecture is inherently agnostic to the supervision source, it naturally accommodates the task reward already available in any RL setting. \starwm{}-R adds a reward prediction head that 
directs the reward gradient to guide the spatial mask, matching or exceeding R2-Dreamer on the tasks where default \starwm{} most strongly underperformed.

  Across the six tasks, \starwm{} and \starwm{}-R are the only methods that achieve high returns under both VH and VHR (Figure~\ref{fig:starwm}, right), and on the combined VH+VHR pool \starwm{}-R attains the highest IQM and lowest optimality gap among all evaluated methods (Appendix~\ref{app:full_results}).

\subsection{What Should Train the Attention?}

\label{sec:exp_ablation}

 \begin{wrapfigure}{r}{0.4\textwidth}
 \vspace{-15pt}
    \centering

    \includegraphics[width=0.37\textwidth]{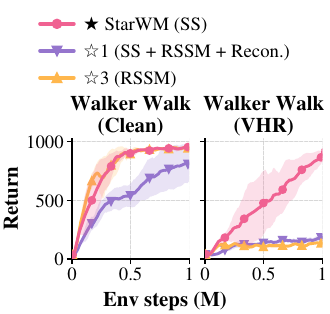}
    \caption{\textbf{Ablation study of sources of the attention signal.} Self-supervised guidance and gradient isolation are jointly necessary.}
    \label{fig:gradient_isolation}
\end{wrapfigure}

The cross-attention parameters can be trained by three signals: self-supervised losses, reconstruction gradients through the decoder, and RSSM gradients through the entity tokens. \starwm{} (\ding{72}) uses stop-gradient barriers to leave the self-supervised losses as the sole signal training the attention parameters. We assess whether this combination is necessary by comparing against two ablations: one (\ding{73}1) that removes the stop-gradients so all three sources reach the attention, and one  (\ding{73}3) that removes the self-supervised loss and makes RSSM gradients the only signal. On NoisyShape, all three configurations yield near-optimal shape retention: the object easiest to reconstruct (the geometric shape) coincides with the action-conditional one, so reconstruction-based gradients guide attention to the correct target without further supervision. On DMC with video distractors, this alignment breaks down and self-supervised guidance becomes critical. Figure~\ref{fig:gradient_isolation} shows that both ablations collapse under VHR while \starwm{} maintains performance; the two design choices are jointly necessary because they together restrict the attention signal to a dynamics-grounded source. Full ablations are in Appendix~\ref{app:gradient_isolation}.

\subsection{When Does Self-Supervised Routing Suffice?}
\label{sec:exp_boundary}

\paragraph{What each signal captures.}
The two default signals select different content. Inverse dynamics points
attention to regions whose changes are explained by the agent's action, an
unambiguous target: on NoisyShape, it alone suffices for near-optimal routing.
Temporal contrastive learning only asks consecutive frames to be more similar
than frames sampled from the batch, which mostly come from other episodes. It
therefore favors any content that stays consistent within an episode and
differs across episodes, and on its own it fails to single out the agent on
NoisyShape, consistent with~\citet{schneider2021contrastive}. On DMC,
contrastive learning helps under VHR, where the agent is the only such content,
but is partially co-opted under VH, where the sequential video qualifies as
well (Appendix~\ref{app:ss_ablation}).

\paragraph{Passive task-relevant content.}
Some task-relevant content is not driven by the agent's actions: a static
goal, or an object that moves only passively. Inverse dynamics gives no signal
for it, so whether it is retained depends on the contrastive signal alone. We
test this with a NoisyShape variant containing a passive object whose color
and position stay fixed within an episode, DMC Reacher (static target), and
Ball-in-cup (passively moving ball) (Appendix~\ref{app:passive}).

\begin{table}[htbp!]
  \caption{\textbf{When default routing retains passive task-relevant content.}
  Retained if the content is static within an episode and differs across
  episodes, unless a coherent video background already plays that role. Reward
  routing (\starwm{}-R) fixes the failure cases. n/a: reward-free environment.}
  \label{tab:boundary}
  \centering
  \small
  \setlength{\tabcolsep}{4pt}
  \begin{tabular}{@{}ll|cc|cc@{}}
    \toprule
    Passive content & Background
      & \makecell{Static within\\episode?}
      & \makecell{Coherent\\video?}
      & \starwm{} & \starwm{}-R \\
    \midrule
    \multirow{2}{*}{Object position ($R^2$)}
      & random     & \ding{51} & \ding{55} & \textbf{0.80} & -- \\
      & sequential & \ding{51} & \ding{51} & 0.13          & -- \\
    \midrule
    \multirow{2}{*}{Reacher target (return)}
      & VHR & \ding{51} & \ding{55} & \textbf{690} & 657 \\
      & VH  & \ding{51} & \ding{51} & 391          & \textbf{789} \\
    \midrule
    \multirow{2}{*}{Ball-in-cup ball (return)}
      & VHR & \ding{55} & \ding{55} & 98  & \textbf{303} \\
      & VH  & \ding{55} & \ding{51} & 114 & \textbf{304} \\
    \bottomrule
  \end{tabular}
\end{table}

Table~\ref{tab:boundary} follows directly from what the contrastive signal
favors. A static passive object stays fixed within an episode and differs
across episodes, so it is retained, but only when nothing else fits better.
A sequentially played video fits equally well, takes over the contrastive
signal, and the passive object is dropped. A passively moving object, like the
ball, is not consistent within an episode, so it is lost under both
backgrounds.

This is a limitation of self-supervision in general: from observations alone,
a model cannot tell whether predictable but uncontrollable content matters for
the task. Reward provides exactly this missing information. Because the
routing module accepts any training signal, adding reward prediction
(\starwm{}-R) requires no architectural change and recovers all failure cases
in Table~\ref{tab:boundary}.
\section{Related Work}
\label{sec:related}

\paragraph{World models and distraction-robust representation learning.}
Visual world models divide broadly into reconstruction-based approaches (e.g., the Dreamer family~\citep{hafner2020dream, hafner2021dreamerv2, hafner2025mastering}) and reconstruction-free methods (e.g., TD-MPC~\citep{hansen2022tdmpc, hansen2024tdmpc2}). While reconstruction provides dense visual grounding, it suffers from reconstruction bias toward dominant pixel areas; conversely, reconstruction-free methods eliminate this bias but discard pixel-level grounding. \starwm{} bridges this divide by retaining dense reconstruction for content supervision while redirecting capacity away from distractors via self-supervised attention routing.

To tackle visual distractions, prior reconstruction-based methods either substitute reconstruction with auxiliary objectives (DreamerPro~\citep{deng2022dreamerpro}, R2-Dreamer~\citep{morihira2026rdreamer}) or decompose a shared latent space (TIA~\citep{fu2021tia}, DenoisedMDP~\citep{wang2022denoised}, Iso-Dream~\citep{pan2022iso}), but the latter introduces gradient tension across objectives. Spatial approaches localize relevant regions via unsupervised keypoints~\citep{wang2021vai}, simulator masks~\citep{kim2025make}, or SAM-based foundation models~\citep{wang2023generalizable}, yet rely on external supervision. SMG~\citep{zhang2024focus} uses a dual-branch structure to isolate task-relevant reconstruction, but it is model-free (built on DrQ-v2~\citep{yarats2022mastering}) and requires per-task image augmentations. In contrast, \starwm{} derives its spatial routing endogenously from environment dynamics, requiring neither external priors nor task-specific augmentations while preserving dense reconstruction for attended content.
\paragraph{Object-centric learning and cross-attention.}
Slot Attention~\citep{locatello2020object} finds objects using repeated attention steps guided by image reconstruction, and it rebuilds scenes using spatial broadcast decoding. DETR~\citep{carion2020detr} finds objects using learned query vectors in a single attention step. \starwm{} combines this single-step attention with spatial broadcast decoding. In Slot Attention, pixel reconstruction guides the attention weights, which easily leads the model to focus on large or complex backgrounds. In contrast, \starwm{} trains its attention using self-supervised dynamics and blocks all gradients from reconstruction. In addition, using a single attention step makes it easy to add stop-gradient barriers between attention routing and feature extraction, which repeated attention steps cannot do.

\paragraph{Self-supervised methods for RL.}
Prior self-supervised RL methods often use inverse dynamics~\citep{pathak2017curiosity}, contrastive learning~\citep{laskin2020curl, oord2018infonce}, or predictive objectives~\citep{schwarzer2021data} to learn whole-image representations, which drops spatial layout. In contrast, \starwm{} uses these dynamics signals to train a spatial attention module that routes information, while keeping image reconstruction to supervise the visual details of attended parts. The attention gating mechanism is also supervision-agnostic. We default to inverse dynamics and contrastive learning for reward-free applicability, but task-specific signals such as reward prediction (\S\ref{sec:exp_dmcgb}) can be added. 

\section{Conclusion}
\label{sec:conclusion}
A central challenge in robust world model learning is the tension between reconstruction-based objectives, which provide pixel-level grounding to the observation space but overemphasize irrelevant detail, and reconstruction-free objectives, which ignore noise but can lose key environment information. We introduced \starwm{}, which unites the strengths of both approaches. By using learned attention to guide reconstruction only to relevant regions, \starwm{} stops background distractors from entering the latent dynamics model. This prevents the model from wasting capacity on large irrelevant areas, enabling accurate predictions over long horizons. Our default setup uses inverse dynamics and contrastive learning to find controllable and predictable objects without any task rewards, clearly outperforming reconstruction-based baselines under visual distractions. Moreover, this attention design is modular and can easily include other signals, such as task reward in \starwm{}-R. More broadly, the dual stream decomposition in \starwm{} echoes the classical what-versus-where pathways in visual neuroscience~\citep{MISHKIN1983414}, hinting at the possibility of a more general design principle. 

\paragraph{Limitations.}
Our experiments focus on single-agent continuous control from images. Settings with multiple agents, non-spatial inputs, or real-world domain shifts remain untested, although preliminary tests (Appendix~\ref{app:multi_object}) show that attention queries can separate across multiple objects. Additionally, the $8{\times}8$ attention grid can be too coarse for tiny objects (Appendix~\ref{app:small_objects}). Our default attention signals also have trade-offs: inverse dynamics gives weak guidance for passively moving objects (such as the ball in Ball-in-Cup), static targets under coherent backgrounds (Appendix~\ref{app:passive}), or delayed action effects (Appendix~\ref{app:delay}), while contrastive learning can be fooled by moving video backgrounds. While adding reward prediction fixes these issues in \starwm{}-R, finding a single self-supervised signal that works in every setting remains an open question (Appendix~\ref{app:failure_cartpole}). Finally, keeping an image decoder requires more compute than reconstruction-free methods like R2-Dreamer, though the training cost matches the original DreamerV3 backbone (Appendix~\ref{app:baseline}).

\begin{ack}
The authors acknowledge funding and support from the German Aerospace Center (DLR), the Deutsche Forschungsgemeinschaft (DFG, German Research Foundation) under grant GRK 3012 – 520750254 (kemai.uni-ulm.de), and the state of Baden-Württemberg through bwHPC and DFG grant INST 35/1597-1 FUGG.
\end{ack}

\bibliography{references}
\bibliographystyle{plainnat}

\newpage
\appendix
\addtocontents{toc}{\protect\setcounter{tocdepth}{2}}
\begin{center}
  \Large\textbf{Appendix: Table of Contents}
\end{center}
\vspace{0.5em}
\makeatletter
\@starttoc{toc}
\makeatother
\vspace{2em}
\newpage

\section{Extended Related Work}
\label{app:extended_related}

This section provides an in-depth discussion of how \starwm{} relates to and differs from existing methods across several research axes.

\subsection{Taxonomy of Observation Components}
\label{app:taxonomy}

A visual observation in a typical RL environment contains multiple types of information, which can be classified along three orthogonal axes:

\begin{itemize}
    \item \textbf{Predictable vs.\ unpredictable.} Can the component's future values be predicted from past states and actions? The agent's body dynamics are predictable; randomly sampled background textures are not.
    \item \textbf{Controllable vs.\ uncontrollable.} Is the component's transition function conditionally dependent on the agent's action? The agent's joint angles are controllable; a moving cloud in the background is not.
    \item \textbf{Task-relevant vs.\ task-irrelevant.} Does the component affect the reward function? The ball's position in a catching task is task-relevant; wall texture is not.
\end{itemize}

Different prior methods implicitly target different subsets of this taxonomy:
\begin{itemize}
    \item \textbf{DBC}~\citep{zhang2021learning}: separates task-relevant from task-irrelevant via bisimulation metrics. Requires a meaningful reward signal.
    \item \textbf{TIA}~\citep{fu2021tia}: separates task-relevant from task-irrelevant via reward-prediction adversarial learning. Brittle under sparse rewards and non-transferable across tasks.
    \item \textbf{DenoisedMDP}~\citep{wang2022denoised}: provides a four-way decomposition based on controllability $\times$ reward-relevance via information-theoretic objectives. Operates at the final encoder output without explicit spatial reasoning.
    \item \textbf{Iso-Dream / Iso-Dream++}~\citep{pan2022iso, pan2023isodreampp}: separate controllable from noncontrollable latent transitions using inverse dynamics, then roll out the two branches separately for behavior learning. Like \starwm{}, uses inverse dynamics as the separation signal, but applies it at the latent level rather than the spatial level.
    \item \textbf{HRSSM}~\citep{sun2024hrssm}: combines spatio-temporal cube masking with bisimulation-based similarity to learn compact reward-aware representations, using a hybrid RSSM where a mask branch and an EMA raw branch share history. Operates at the global feature level rather than spatially, and the bisimulation component requires reward signals.

\item \textbf{SeeX}~\citep{huang2024seex}: targets unsupervised exploration with a dual-branch world model that mirrors \starwm{}'s relevant/irrelevant split at the architectural level. Separation is driven by an exogenous reconstruction loss that requires $s^-$ to reconstruct the full observation, exploiting the prior that task-relevant content occupies a small portion of pixels. Unlike \starwm{}, the two branches operate on entire images without spatial routing, and the policy is learned via intrinsic reward maximization rather than task reward.
    \item \textbf{ExoRL}~\citep{efroni2022sample}: formally defines the ExoMDP framework, separating endogenous (action-affected) from exogenous (action-independent) state factors. It provides theoretical regret guarantees but does not address the pixel-level representation challenges in visual RL.
    
\end{itemize}

\starwm{} differs from these methods on two axes. First, the decomposition operates at spatial resolution through explicit cross-attention, rather than at the latent level as in DenoisedMDP and Iso-Dream/Iso-Dream++. Second, \starwm{} is a framework rather than a fixed objective: any differentiable signal applied to the live attention path trains the routing without interfering with reconstruction, by virtue of the stop-gradient isolation. Our default instantiation combines inverse dynamics (controllability axis) with temporal contrastive learning (predictability axis), neither of which requires reward; when reward is available, it can be added as an additional attention-guidance signal along the task-relevance axis (\S\ref{app:reward_extension}). The choice of attention loss thus becomes a modular design decision based on task characteristics, rather than a fixed architectural commitment.

\subsection{Self-Supervised Signals and Their Limitations}
\label{app:selfsup_limits}

Self-supervised signals provide an attractive alternative to reward-based supervision because they are derived from the agent's interaction with the environment without requiring task labels. The most popular such signals fall into three families: state-only contrastive learning (e.g., CURL~\citep{laskin2020curl}), forward-dynamics prediction (e.g., SPR~\citep{schwarzer2021data}), and inverse-dynamics prediction~\citep{pathak2017curiosity}.

\citet{rakelly2021mutual} analyze these three families theoretically and show that none of them is a reliable representation learner for control alone. State-only contrastive objectives can yield representations that alias states with different optimal actions. Inverse-dynamics objectives can discard task-relevant state information that is not necessary for predicting actions. Forward-dynamics objectives are the most robust of the three, but retain any state information that is temporally correlated.

These results motivate \starwm{}'s design. Rather than rely on any single self-supervised objective as the primary representation learner, we use them as attention guidance signals: inverse dynamics steers attention toward action-conditional regions, and contrastive learning provides complementary temporal supervision in transitions where inverse dynamics is silent. The visual content of attended regions is supervised separately by reconstruction through the dual-stream decoder, and stop-gradient isolation prevents reconstruction gradients from corrupting the attention parameters or vice versa. In this way, each signal operates within the regime where its inductive bias is robust, and no single signal carries the full burden of representation learning.

\subsection{DreamerV3's Built-in Noise Handling}
\label{app:dreamerv3_noise}

DreamerV3 already incorporates several features designed to handle stochastic environments:
\begin{itemize}
    \item \textbf{Dual latent variables.} The RSSM maintains both a deterministic recurrent state $h_t$ (encoding history) and a stochastic categorical state $z_t$ (capturing single-step uncertainty). In principle, the stochastic component can absorb unpredictable variation, while the deterministic component tracks persistent dynamics.
    \item \textbf{KL balancing and free nats.} KL balancing prevents posterior collapse by asymmetrically scaling the KL divergence between the posterior and prior. The free nats mechanism sets a minimum KL threshold, allowing the model to maintain a minimum level of stochasticity in $z_t$ without being penalized.
    \item \textbf{Symlog scaling.} Symlog transformations handle varying signal magnitudes, improving stability across diverse domains.
\end{itemize}

These mechanisms are effective for latent-level noise handling: the stochastic $z_t$ can represent uncertainty over globally unpredictable factors, and the prior $p(z_t|h_t)$ can learn to ``ignore'' unpredictable components whose KL cost exceeds their reconstruction benefit. However, they operate after the encoder has already compressed the full $64 \times 64$ observation into a single vector. Three core vulnerabilities remain upstream:

\begin{itemize}
    \item \textbf{The encoder bottleneck.} The dense image decoder requires the encoder to faithfully encode all visual content. The RSSM's noise-handling mechanisms cannot undo this upstream encoding pressure; they can only attempt to filter the already-contaminated latent.
    \item \textbf{Spatially entangled representations.} Without spatial attention, attended and background features are entangled in the same latent vector. The stochastic $z_t$ lacks the expressivity to cleanly separate spatially localized entities from distributed background patterns. Even with free nats, the KL budget is shared between genuine action-induced uncertainty and background stochasticity.
    \item \textbf{Imagination rollouts.} During imagination, the world model generates $z_t$ from the prior $p(z_t|h_t)$. If the posterior $z_t$ was contaminated by background information during encoding, the prior must also model this contamination to match. This leads to hallucinated background dynamics in imagined trajectories, directly degrading policy quality.
\end{itemize}

\starwm{} addresses this gap by intervening before the RSSM: the cross-attention module spatially separates attended entities from backgrounds at the feature map level, feeding only objective-relevant entity features into the RSSM. The dual-stream decoder removes the reconstruction pressure on the RSSM by handling background reconstruction through a separate VAE pathway that is discarded during imagination. The RSSM's built-in noise handling mechanisms can then focus entirely on the dynamics of endogenous entities, rather than wasting capacity on filtering background contamination.

\subsection{Comparison with Object-Centric and Slot-Based Methods}
\label{app:slot_comparison}
\paragraph{Object-centric methods.}
Among object-centric methods, Slot Attention~\citep{locatello2020object} is the most closely related line to \starwm{} in mechanism, sharing the use of attention to extract entity-level representations from spatial feature maps. However, the two approaches differ fundamentally in what trains the attention:

\begin{center}
\begin{tabular}{lcc}
\toprule
 & \textbf{Slot Attention} & \textbf{\starwm{}} \\
\midrule
Attention supervision & Reconstruction & Self-supervised Signals \\
Attention refinement & Iterative & Single-pass \\
Slot competition & Softmax over slots & Independent softmax per query \\
Gradient from recon. & Flows through attention & Blocked by stop-gradient \\
\bottomrule
\end{tabular}
\end{center}

For reconstruction-supervised attention, the mechanism is trained to minimize reconstruction error, which is proportional to the number of pixels covered. This incentivizes slots to attend to visually complex, spatially dominant regions, such as the distractor backgrounds that should be ignored. Additionally, Slot Attention's iterative refinement couples the attention weights to the decoder, making it infeasible to apply a simple stop-gradient fix without breaking the iterative convergence.

\starwm{}'s self-supervised attention training is immune to this bias: inverse dynamics provides gradients only for action-conditional regions, while the stop-gradient ensures that reconstruction gradients never reach the attention parameters. This makes the attention mechanism robust to arbitrary background complexity without requiring any modification to the loss functions.

\paragraph{Object-centric world models.}
The broader landscape of object-centric world models includes both slot-attention-based methods (e.g., SlotFormer~\citep{wu2023slotformer}, SLATE~\citep{singh2022slate}) and methods relying on external segmentation supervision (e.g., FOCUS~\citep{ferraro2023focus}, which uses ground-truth masks from simulators or SAM). These methods aim at scene decomposition into discrete objects for prediction or control, typically in environments with well-defined object boundaries. \starwm{} addresses a different regime: distractor robustness in visual control, where the failure mode is the reconstruction trap rather than object decomposition. It is not designed for full multi-object scene parsing; nevertheless, our two-object experiment (Appendix~\ref{app:multi_object}) shows that attention queries spontaneously partition across relevant entities even in this distractor regime, suggesting compatibility with multi-object settings as a future direction.

\section{Method \& Implementation Details}
\subsection{Algorithm}
\label{app:algorithm}
\begin{figure}[htbp!]
    \centering
    \includegraphics[width=0.82\textwidth]{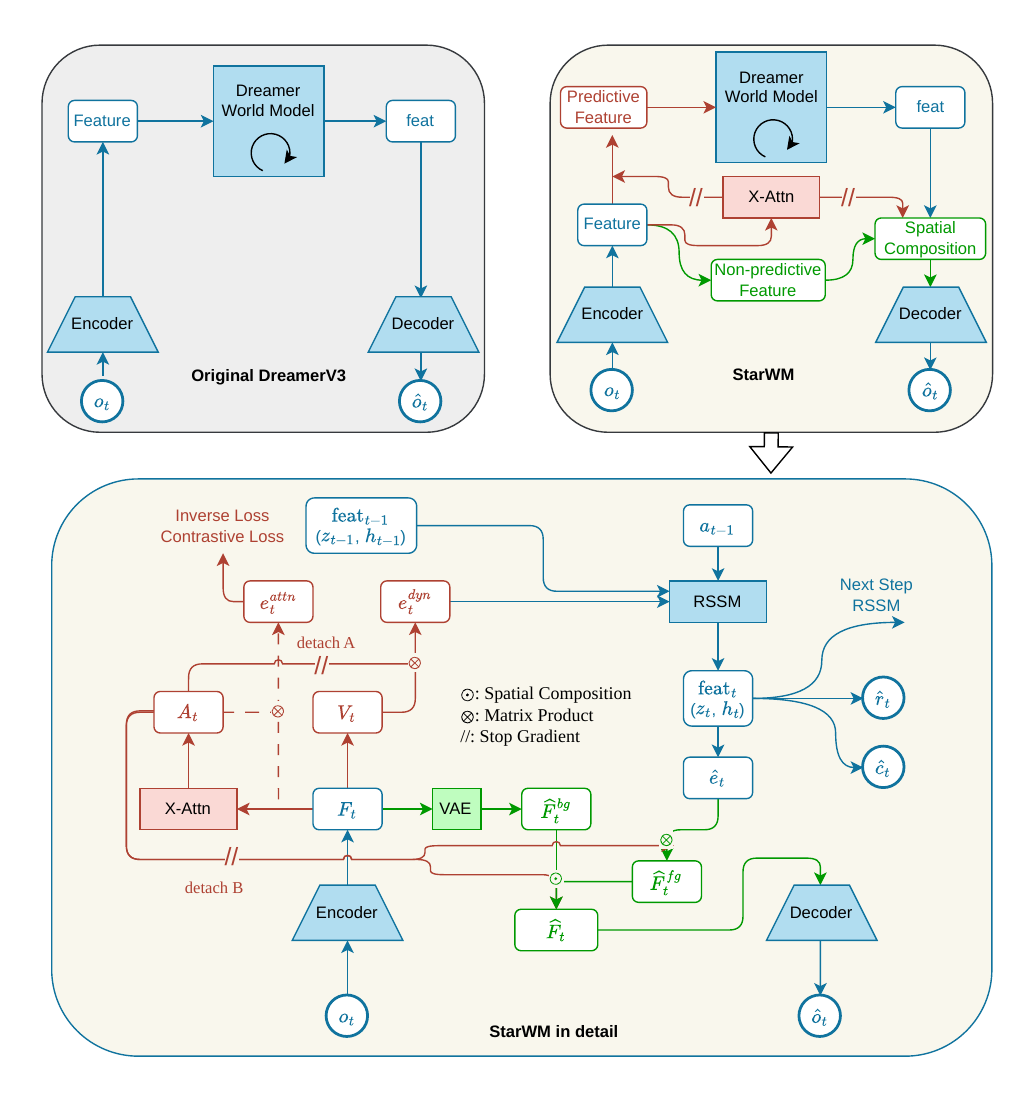}
\caption{\textbf{\starwm{} architecture.} Top: comparison with DreamerV3. \starwm{} splits the encoder output via cross-attention into an attended stream (entering the RSSM) and a residual stream (auxiliary VAE, discarded during imagination). Bottom: detailed architecture. The cross-attention module (red) extracts entity tokens guided by self-supervised losses; the dual-stream decoder (green) composes attended foreground content with residual background content for reconstruction. Blue components are inherited from DreamerV3. Stop-gradient barriers (//) isolate attention parameters from reconstruction gradients.}

    \label{fig:architecture}
\end{figure}
Algorithm~\ref{alg:starwm} presents the complete training procedure for one 
batch. Below we detail implementation choices omitted from the main text for 
brevity.
\begin{algorithm}[ht]
\caption{\starwm{} World Model Training (one batch)}
\label{alg:starwm}
\begin{algorithmic}[1]
\REQUIRE Batch of sequences $\{(o_t, a_t, r_t, c_t)\}_{t=1}^{T}$
\STATE \textbf{// Encoder interception}
\STATE $F^8_t \leftarrow \mathrm{Encoder}_{\text{layers}\,1\text{-}3}(o_t)$ \hfill $\triangleright$ Intermediate 8$\times$8 features
\STATE $F_t \leftarrow \mathrm{flatten}(\mathrm{Conv}_{1\times 1}(F^8_t))$ \hfill $\triangleright$ Project to $L \times C $
\STATE $F^4_t \leftarrow \mathrm{Encoder}_{\text{layer}\,4}(F^8_t)$ \hfill $\triangleright$ Full encoder output at 4$\times$4
\STATE
\STATE \textbf{// Cross-attention spatial routing}
\STATE $K_t \leftarrow F_t +E_\mathrm{pos}$ \hfill $\triangleright$ Keys with positional encoding
\STATE $V_t \leftarrow [F_t;\; \mathrm{E_\mathrm{coord}} ]$ \hfill $\triangleright$ Values with coordinate grid
\STATE $A_t \leftarrow \mathrm{MultiHeadAttn}(Q, K_t)$ \hfill $\triangleright$ Attention weights $\in \mathbb{R}^{N \times L}$
\STATE $e^{\text{attn}}_t \leftarrow \mathrm{flatten}(A_t  F_t)$ \hfill $\triangleright$ Entity features (live $A$)
\STATE
\STATE \textbf{// Self-supervised attention training (gradient flows through $A$)}
\STATE $\mathcal{L}_{\text{inv}} = \ell\!\big(\phi_\mathrm{inv}([e_t^{\text{attn}};\, e_{t+1}^{\text{attn}}]),\; a_t\big)$
\STATE $ \mathcal{L}_{\text{ctr}} = -\log\sigma\!\left(\tfrac{\phi_\mathrm{ctr}(e_t^{\text{attn}})^\top \phi_\mathrm{ctr}(e_{t+1}^{\text{attn}})}{\tau}\right) - \log\sigma\!\left(-\tfrac{\phi_\mathrm{ctr}(e_t^{\text{attn}})^\top \phi_\mathrm{ctr}(\tilde e^{\text{attn}})}{\tau}\right)$
\STATE
\STATE \textbf{// Stop-gradient decoupled pathway (gradient blocked at $\bar{A}$)}
\STATE $\bar{A}_t \leftarrow \mathrm{sg}(A_t)$ \hfill $\triangleright$ Block gradients to attention params
\STATE $e_t^{\text{dyn}} \leftarrow \phi_{\mathrm{dyn}}(\bar{A_t}  V_t)$ \hfill $\triangleright$ RSSM input
\STATE $\mathrm{feat}_t$ ($[h_t,z_t]$), $\leftarrow \mathrm{RSSM}$ \hfill $\triangleright$ Latent dynamics

\STATE $\mathcal{L}_{\text{KL}}^{\text{RSSM}} \leftarrow \mathrm{RSSM}(e_t^{\text{dyn}}, a_{t-1})$ \hfill $\triangleright$
\STATE $\mathcal{L}_{\text{reward}} \leftarrow -\ln p_\theta(r_t \mid \mathrm{feat}_t)$
\STATE $\mathcal{L}_{\text{cont}} \leftarrow -\ln p_\theta(c_t \mid \mathrm{feat}_t)$
\STATE
\STATE \textbf{// Dual-stream decoder}
\STATE $\hat{F}_t^{\text{fg}} \leftarrow \bar{A}_t^\top \cdot \phi_{\text{fg}}(\mathrm{feat}_t)$ \hfill $\triangleright$ attended: inverse broadcast
\STATE $z_t^{\text{bg}} \sim q_\psi(z_t^{\text{bg}} \mid \mathrm{flatten}(F^4_t))$ \hfill $\triangleright$ Background: VAE encoding
\STATE $\hat{F}_t^{\text{bg}} \leftarrow \phi_{\text{bg}}(z_t^{\text{bg}})$
\STATE $M \leftarrow \mathrm{clamp}(\textstyle\sum_n \bar{A}_{t,n},\, 0,\, 1)$ \hfill $\triangleright$ Coverage mask
\STATE $\hat{o}_t \leftarrow \mathrm{Decode}(\hat{F}_t^{\text{fg}} \odot M + \hat{F}_t^{\text{bg}} \odot (1{-}M))$
\STATE $\mathcal{L}_{\text{img}} \leftarrow \| \hat{o}_t - o_t \|^2$

\STATE
\STATE \textbf{// Joint optimization (stop-gradient prevents conflicting gradients)}
\STATE $
    \mathcal{L} = \underbrace{\mathcal{L}_{\text{img}} + \mathcal{L}_{\text{reward}} + \mathcal{L}_{\text{cont}} + \mathcal{L}_{\text{KL}}^{\text{RSSM}}}_{\text{Dreamer backbone}} + \underbrace{\lambda_{\text{inv}}\,\mathcal{L}_{\text{inv}} + \lambda_{\text{ctr}}\,\mathcal{L}_{\text{ctr}}}_{\text{X-Attention guidance}} + \underbrace{\beta_{\text{bg}}\,\mathcal{L}_{\text{KL}}^{\text{bg}}}_{\text{Residual regularization}}.
$
\STATE Update all parameters with $\nabla\mathcal{L}$
\end{algorithmic}
\end{algorithm}
\paragraph{Encoder interception.}
Standard DreamerV3 compresses a $64{\times}64$ image through four convolutional stages to a $4{\times}4$ feature map ($16$ spatial tokens) before the RSSM. \starwm{} intercepts the encoder after its third convolutional stage, retaining an intermediate $8{\times}8$ feature map $F_8$ with $128$ channels ($L{=}64$ spatial tokens). A $1{\times}1$ convolution projects $F_8$ to $C=256$ channels. The fourth stage continues normally to produce $F_4$ ($256$ channels, $4{\times}4$), which is flattened to a $4096$-dimensional vector for the background VAE stream (\S\ref{sec:what}).
\paragraph{Multi-head attention aggregation.}
\label{app:mha_detail}
The cross-attention module (\S\ref{sec:where}) uses $H{=}4$ heads with 
head dimension $C/H{=}64$. Query and key projections $W_Q, W_K$ operate 
per-head; attention logits are computed independently per head and are
softmaxed. The final attention map $A_t \in \mathbb{R}^{N \times L}$ is the 
arithmetic mean across heads. Entity tokens are then extracted by applying 
this averaged map to the unprojected values, i.e.\ $e_t^{dyn} = \bar{A}_t \cdot V_t$. 
\paragraph{Image loss.}
\starwm{} uses per-pixel MSE loss summed over spatial dimensions, replacing 
DreamerV3's symlog-discretized distribution head for the image decoder. This 
simplification is sufficient because the dual-stream architecture handles 
signal magnitude differences between foreground and background through 
explicit spatial routing rather than output normalization.

\paragraph{Inverse dynamics for continuous vs.\ discrete actions.}
The inverse dynamics loss $\mathcal{L}_{\mathrm{inv}}$ (Eq.~\ref{eq:inv}) 
uses cross-entropy for discrete action spaces (NoisyShape, 4 actions) and 
mean squared error for continuous action spaces (DMC).

\paragraph{Contrastive negative sampling.}
The temporal contrastive loss (Eq.~\ref{eq:ctr}) uses binary InfoNCE with 
one positive (the temporally adjacent frame) and one negative (a randomly 
sampled frame from the same batch). This is computationally lightweight and 
empirically sufficient given that the primary attention guidance comes from 
inverse dynamics.

\paragraph{Background VAE KL annealing.}
\label{app:annealing}
The background KL scale $\beta_{\mathrm{bg}}$ is annealed linearly from 
$\beta_{\mathrm{bg}}^{\mathrm{init}} = 5.0$ to 
$\beta_{\mathrm{bg}}^{\mathrm{final}} = 0.5$ over the first $15{,}000$ 
training steps. The high initial value encourages the background VAE to 
produce near-prior samples early in training, preventing it from encoding 
foreground content before the attention module has learned to localize 
entities. After annealing, the lower $\beta_{\mathrm{bg}}$ permits the 
background stream to reconstruct richer background detail without competing 
with the dynamics pathway.

\subsection{Hyperparameters}
\label{app:hyperparams}

Table~\ref{tab:hyperparams} lists the key hyperparameters of \starwm{} used throughout all experiments.

\begin{table}[ht]
  \caption{Hyperparameters of \starwm{}.}
  \label{tab:hyperparams}
  \centering
  \small
  \begin{tabular}{llc}
    \toprule
    \textbf{Component} & \textbf{Hyperparameter} & \textbf{Value} \\
    \midrule
    Cross-attention & Number of queries $N$ & 4 \\
    & Number of heads $H$ & 4 \\
    & Head dimension $d/H$ & 64 \\
    & Projected channels $d$ & 256 \\
    & Intercept resolution & $8 \times 8$ \\
    & Positional encoding & 2D sinusoidal \\
    \midrule
    Background VAE & Latent dimension $D_\text{bg}$ & 128 \\
    & KL free bits $\kappa$ & 1.0 \\
    & KL scale $\beta_\text{bg}$ (final) & 0.5 \\
    & KL scale $\beta_\text{bg}$ (initial) & 5.0 \\
    & Anneal steps $T_{\text{anneal}}$ & 15{,}000 \\
    \midrule
    Self-supervised & Inverse dynamics scale $\lambda_\text{inv}$ & 1.0 \\
    & Contrastive scale $\lambda_\text{ctr}$ & 1.0 \\
    & Contrastive temperature $\tau$ & 0.1 \\
    & Projection dimension $D_p$ & 128 \\
    \midrule
    Decoder & Upsampling stages & 3 ($8 \!\to\! 64$) \\
    & Channel progression & $256 \!\to\! 128 \!\to\! 64 \!\to\! 3$ \\
    & Kernel size & 4 \\
    \midrule
    DreamerV3 backbone & RSSM deterministic units & 512 \\
    & Discrete latents $z$ & 32 classes $\times$ 32 categories \\
    & Encoder CNN depth & 32 (channels: $32 \!\to\! 64 \!\to\! 128 \!\to\! 256$) \\
    & RSSM KL scale $\beta$ & 0.5 \\
    & KL free bits & 1.0 \\
    \bottomrule
  \end{tabular}
\end{table}
\subsection{Implementation of Baseline Algorithms and  Compute Resources}
\label{app:baseline}
We compare our method against three representative model-based reinforcement learning baselines: DreamerV3 \citep{hafner2025mastering}, R2-Dreamer \citep{morihira2026rdreamer}, DenoisedMDP \citep{wang2022denoised}, and Iso-Dream++ \citep{pan2023isodreampp}. Original implementation of DreamerV3 is in JAX, and We use the PyTorch reimplementation on GitHub[NM512/dreamerv3-torch]. For other baselines, we use the official open-source implementation and adopt the recommended default hyperparameters. To ensure that all methods are evaluated under identical observation conditions, we make minimal modifications to integrate a unified Video Hard background distractor wrapper, while leaving the underlying agent architecture, optimization procedure, and learning hyperparameters untouched. For dmc tasks, We use the dmc\_vision configuration as provided in the official codebase for each baseline.

All baselines and our method are run within the same conda environment, on a single NVIDIA A100 / H100 GPU per training run. Each training run uses three random seeds; for each task–method combination, the three seeds are packed onto a single GPU and trained concurrently. Training budgets are set to 3M environment steps for the VH variant and 1M environment steps for the VHR variant. Table~\ref{tab:compute_per_alg_mode} reports the resulting average GPU-hours per task. The reported wall-clock numbers should be interpreted as approximat as the same GPU model exhibits speed variation depending on cluster load and concurrent jobs, and A100/H100 differ in raw throughput. 
\begin{table}[h]
\centering
\caption{Average GPU-hours per task on a single NVIDIA H100/A100 (one training run packs 3 random seeds onto the same GPU). Numbers are total wall-clock time to reach the target step budget (1M for Clean and VHR; 3M for VH).}
\label{tab:compute_per_alg_mode}
\begin{tabular}{lccc}
\toprule
Algorithm & Clean (1M) & VH (3M) & VHR (1M) \\
\midrule
StarWM (ours)    &  35 & 94 & 34 \\
Orig DreamerV3   & 30 & 104 & 37 \\
Iso-Dream++      & --- & 103 & 37 \\
DenoisedMDP      & --- & 37 & 15 \\
R2-Dreamer       & --- & 26 & 10 \\

\bottomrule
\end{tabular}
\end{table}

\starwm{} runs at comparable speed to other reconstruction-based world models (DreamerV3, Iso-Dream++) and is slower than R2-Dreamer, which forgoes the image decoder. This is consistent with our limitations discussion: retaining reconstruction provides visual grounding at the cost of higher per-step compute. The cross-attention and self-supervised auxiliary heads add minor overhead relative to the DreamerV3 backbone \starwm{} builds on.

\section{Environments \& Evaluation Protocols}
\label{app:env_design}

\subsection{Shape Environments}
\label{app:shapeenv}

The Shape family of environments provides a controlled and interpretable testbed for analyzing how world models encode and retain objective-relevant attributes under visual distractions.

\begin{itemize}
  \item \textbf{Shape and Scale:} There are four geometric shapes (circle, triangle, square, and pentagon) arranged in a fixed cycle. Each shape appears in one of three sizes: small, medium, or large. When the agent applies the \textit{forward shape} action, the object increases in size; once it reaches the largest size, it transitions to the next shape in its smallest form. The \textit{backward shape} action reverses this progression.
  \item \textbf{Color:} The object color cycles through red $\rightarrow$ green $\rightarrow$ blue $\rightarrow$ black and vice versa, controlled by the other two discrete actions (\textit{forward color} and \textit{backward color}).
  \item \textbf{Position:} The spatial location of the object is randomized independently at each time step and does not respond to agent actions.
  \item \textbf{Noise:}  Noisy backgounds.
\end{itemize}

The agent's action space is thus composed of four discrete actions, corresponding to the forward/backward transitions of shape and color. 

\paragraph{Shape environment details.}

The observations in the Shape environment are rendered as images with an initial resolution of $64 \times 64$ pixels and $3$ channels (RGB).

The position of the geometric shape in each image is randomized using a uniform distribution, constrained to ensure that the shape is fully contained within the image boundaries. Every shape, whether a circle, triangle, square, or pentagon, uses the same radius value for each size, meaning small, medium, and large mean the same thing across all of them. However, due to their differing geometries, the resulting areas vary across shape types.

The mapping between shape type, size, and the corresponding radius and area is summarized in Table~\ref{tab:shape_sizes}. 
\begin{figure}[t]
    \centering
    \includegraphics[width=\textwidth]{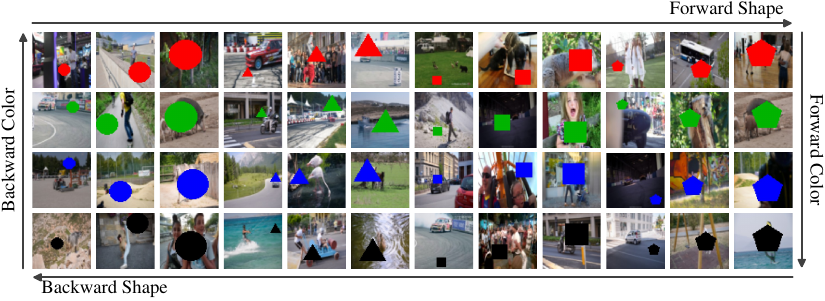}
\caption{\textbf{\ref{app:shapeenv}: Illustration of NoisyShape Environments.} }

    \label{fig:noisyshapeenv}
\end{figure}

	\begin{table}[!htbp]
		\caption{Parameters for shapes. }
  \label{tab:shape_sizes}
		\begin{center}
		\begin{tabular}{c|c|c|c}
		\toprule
		 & Small, $r=6.4$ & Medium, $r=11.52$ & Large, $r=16.64$\\
		\midrule
  Circle &  $S=129$ ($3.14\%$)& $S=417$ ($10.18\%$) & $S=870$ ($21.24\%$)\\
    \midrule
    Triangle &$S=53$ ($1.30\%$) &  $S=172$ ($4.21\%$) &  $S=360$ ($8.78\%$)\\
    \midrule
    Square &  $S=82$ ($2.00\%$)&  $S=265$ ($6.48\%$)&  $S=554$ ($13.52\%$)\\
    \midrule
    Pentagon &  $S=97$ ($2.38\%$)&   $S=316$ ($7.70\%$)&  $S=658$ ($16.07\%$)\\
		 \bottomrule
		\end{tabular}
		\end{center}
		\end{table}

\paragraph{Features of the environment.}
The Shape environment contains diverse observational features that differ in predictability and semantic relevance. Noise is entirely stochastic and neither predictable nor reconstructable, acting as irrelevant variation that should ideally be ignored by the representation. Position, though fully reconstructable from the observation, is unpredictable due to random sampling at each step. It is often retained by reconstruction-based methods despite being semantically uninformative. Shape and size form a coupled, action-dependent dynamic; size changes more rapidly and triggers shape transitions, making both features highly predictable and essential for modeling controllable structure. However, note that only size information is essential to predict actions between neighbor observations. In contrast, color evolves independently of shape and size. Its small discrete state space and weak temporal continuity make it harder to reliably model using temporal objectives such as contrastive learning. This decomposition illustrates the challenges in disentangling useful dynamics from distractors and guides the design of robust representation objectives.

\paragraph{Chance level calculation.}
For categorical attributes, chance accuracy follows directly from the number of classes: $1/4$ for color (4 colors), $1/4$ for shape (4 shapes), and $1/3$ for size (3 sizes).

For position, the target object's normalized coordinates $(x, y)$ are sampled independently and uniformly within $[m, 1-m]^2$, where $m$ is the object-size-dependent margin that prevents the object from extending outside the frame. Since both coordinates are symmetrically distributed around $0.5$, the optimal constant predictor (in MSE) is $\hat{x} = \hat{y} = 0.5$, and its expected squared error equals the sum of the variances of $x$ and $y$:
\begin{equation}
    E\left[(x - 0.5)^2 + (y - 0.5)^2\right] = 2 \cdot \mathrm{Var}(x)  = \frac{(1-2m)^2}{6}.
\end{equation}
The margin $m$ is determined by the object's radius (Table~\ref{tab:shape_sizes}): $m = r / 64 \in \{0.10, 0.18, 0.26\}$ for small, medium, and large sizes respectively. Since the environment cycles through sizes deterministically with equal long-run frequency $1/3$, the chance-level position MSE is
\begin{equation}
    \frac{1}{3} \sum_{m \in \{0.10, 0.18, 0.26\}} \frac{(1-2m)^2}{6} \approx 0.0711.
\end{equation}
We use this value as the dashed chance line in Figure~\ref{fig:probe_curves}.

\paragraph{Design rationale.}
The object occupies $2\%\sim 20\%$ of pixels while the background covers $80\%\sim 98\%$, creating a direct test of pixel-area bias. Rewards are uniform random noise, ensuring that no reward signal can guide representation learning. This forces the model to rely entirely on its inductive biases and training objectives for entity identification.

\paragraph{Variants.}
We have four background conditions:
\begin{itemize}
  \item \textbf{Shape Env} (clean): Solid white background. Position is randomized each step. This isolates the dynamics model's ability to encode shape under minimal distraction.
  \item \textbf{Shape Env-Fixed} (clean, fixed position): shape at a fixed screen position on a plain background. This baseline removes \emph{all} sources of variation except the objective-relevant attributes (color, shape, size), establishing that DreamerV3 can perfectly encode these attributes when no confounding information competes for representational capacity.
  \item \textbf{Noisy Shape Env} (random noise): Each step independently samples a DAVIS~\citep{perazzi2016benchmark} video frame as the background ($64 \times 64$, drawn from all 6208 frames of the 90 DAVIS 2017 train+val sequences). Both position and background are resampled every step, creating maximal stochasticity.
  \item \textbf{Noisy Shape Sequential Env} (sequential noise): Similar to Noisy Shape Env, but the background video frames advance sequentially within each episode (one frame per step from a DAVIS video) rather than being independently sampled. This creates a more realistic but also more challenging distraction pattern: the background has temporal autocorrelation, which may cause the contrastive loss to reward encoding background dynamics.
  \item \textbf{Noisy Shape Env with Uniform Motion} (random noise, positions are predictable): The shape moves in uniform linear motion: at each step, position advances by a fixed velocity vector, reflecting off the image boundaries on contact(a margin of 0.05, i.e.\ positions stay in $[0.05, 0.95]^2$). Initial position and velocity are sampled once per episode and held constant thereafter. All other attributes (color, shape, size, action space, reward noise) follow the NoisyShape setup. Background frames are sampled from a fixed pool of 20 DAVIS frames (the first 20 frames in sorted order, all from a single sequence) and switched independently at each step, providing a clean test of objective-irrelevant signal suppression while  position becomes a dynamically predictable attribute. See Appendix~\ref{app:predictable_position} for results and discussion.
\end{itemize}
\subsection{Probe Analysis Methodology for NoisyShape}
\label{app:probe_method}

The diagnostic probe analysis in \S\ref{sec:exp_diagnostic} evaluates how well the RSSM latent state encodes ground-truth environment attributes. We describe the data collection, probe architecture, and training protocol below.

\paragraph{Data collection.}
We collect trajectories by running a random policy in the target environment. For each experiment, we gather 500 episodes (each 200 steps) for training the probe classifiers, and a separate set of 200 episodes for evaluating imagined rollouts. Both sets are collected once and cached across all models to ensure fair comparison. Actions are sampled uniformly from the discrete action space (4 actions).

\paragraph{Feature extraction.}
For each model, we encode the collected trajectories step-by-step through the frozen world model's encoder and RSSM. At each timestep, we extract the RSSM feature vector $\mathrm{feat}_t = [\,h_t;\; z_t\,] \in \mathbb{R}^{1536}$ (512 deterministic + $32 \times 32$ stochastic). For \starwm{}, encoding follows the cross-attention pathway: the encoder produces $F_8$, which is projected by the $1{\times}1$ convolution, pooled via cross-attention, and passed through $\mathrm{proj}_{\mathrm{dyn}}$ before entering the RSSM. For imagined evaluation, we additionally perform open-loop imagination from each encoded state: starting from the posterior latent, we apply $H{=}15$ consecutive \texttt{img\_step} calls with random actions, extracting the imagined feature at each horizon.

\paragraph{Probe architecture.}
Each probe is a 3-layer MLP: $\mathrm{Linear}(1536, 256) \to \mathrm{ReLU} \to \mathrm{Dropout}(0.2) \to \mathrm{Linear}(256, 256) \to \mathrm{ReLU} \to \mathrm{Dropout}(0.2) \to \mathrm{Linear}(256, C)$, where $C$ is the number of classes for classification (4 for color, 4 for shape, 3 for size) or 1 for regression (position).

\paragraph{Training protocol.}
The dataset is split 70/30 into training and test sets (stratified by label for classification). We train with Adam (learning rate $10^{-3}$, weight decay $10^{-4}$), batch size 512, cross-entropy loss for classification and MSE loss for regression, for up to 100 epochs with early stopping (patience~$=$~10 epochs on validation loss). The best checkpoint by validation loss is restored for final evaluation. Test accuracy (classification) or test MSE (regression) is reported.

\paragraph{Imagined label computation.}
Ground-truth labels at imagination horizon $h$ are computed deterministically from the starting labels and the sequence of imagined actions, following the environment's action logic. For NoisyShape environments, actions 0/1 cycle the color backward/forward among 4 options, and actions 2/3 cycle the combined shape-size index backward/forward among 12 options ($4$ shapes $\times$ $3$ sizes). For NoisyShape with uniform motion, position additionally follows linear motion with elastic boundary reflection, independent of actions; position labels at each horizon are computed by simulating this dynamics forward.

\subsection{DeepMind Control Suite Environments}
\label{app:env_design_dmc}

We evaluate on six continuous control tasks from the DeepMind Control Suite~\citep{tassa2018deepmind}, standard in prior distraction-robust RL work~\citep{fu2021tia, wang2022denoised, deng2022dreamerpro}: \texttt{Walker Walk}, \texttt{Walker Run}, \texttt{Cheetah Run}, \texttt{Cartpole Swingup}, \texttt{Ball-in-cup Catch}, and \texttt{Finger Spin}. All tasks use $64 \times 64$ RGB observations and action repeat 2. We train for 1M environment steps under Clean and VHR settings, and 3M steps under the VH setting to allow convergence under temporally coherent distractors.

\paragraph{Clean.}
Standard DMC rendering with the default background. This serves as a regression test to verify that \starwm{}'s additional modules do not degrade performance when no distractions are present.

\paragraph{Video Hard Sequential (VH).}
The background is replaced by a natural video clip from the DMC-GB \emph{video\_hard} set that plays sequentially within each episode: each simulator step advances the video by one frame, creating a temporally coherent but uncontrollable background distraction. This setting tests robustness to structured, predictable but action-independent visual dynamics. The temporally coherent background is particularly challenging because methods that exploit temporal consistency (e.g., contrastive learning) may inadvertently attend to the background. Rendering details are described below.

\paragraph{Video Hard Random (VHR).}
Same video pool as VH, but each step independently samples a random frame of the current episode's video, destroying all temporal structure. This creates i.i.d.\ high-entropy distractions that are unpredictable by the dynamics model. VHR isolates the effect of spatial visual clutter from temporal distractor dynamics: any method that succeeds under VHR must be filtering distractors based on spatial or action-conditional cues rather than temporal prediction.

\paragraph{Rendering details.}
Our video background settings are based on the DMControl Generalization Benchmark~\citep{hansen2021softda, almuzairee2024recipe}. The original benchmark defines two difficulty levels: \emph{video\_easy}, which retains the ground plane and composites the video behind scene geometry, and \emph{video\_hard}, which removes the ground plane so that the video covers the entire background. We adopt the video\_hard setting for both VH and VHR, as the absence of the ground plane prevents the agent from exploiting a fixed texture as a stable spatial reference. The backgrounds are the 100 \emph{video\_hard} clips distributed with DMC-GB (a subset of RealEstate10K). The first episode of each environment instance starts from a clip chosen by the environment's random seed, and every subsequent episode advances to the next clip of the pool. For VH, the clip is played from its first frame, one frame per simulator step, and loops if the episode exceeds the clip length. For VHR, at each step a frame of the current clip is sampled uniformly. All methods use the same wrapper and video pool. For \texttt{Reacher}, whose top-down camera does not see the backdrop, DAVIS 2017~\citep{perazzi2016benchmark} frames are instead composited behind the arm and target at pixel level (VH: a random clip played sequentially from a random start frame; VHR: a random frame of a random clip at every step).

\section{Full results \& Aggregate Metrics}
\label{app:full_results}
This section provides the full numerical results underlying the main text's 
figures and tables, including standard deviations across seeds.
\subsection{Full results in NoisyShape Environment}
\label{app:full_noisyshape}
Table~\ref{tab:noisyshape_full} reports Probe accuracy on the NoisyShape 
diagnostic environment (\S\ref{sec:exp_diagnostic}) with standard deviations 
across 3 seeds. 
\begin{table}[h]
  \caption{\textbf{\ref{app:full_noisyshape}: Latent probe accuracy on Noisy Shape.} H0 = encoding step; H15 = 15 steps of open-loop imagination. 
    Each ablation isolates one component of \starwm{}.}
  \label{tab:noisyshape_full}
  \centering
  \small
  \setlength{\tabcolsep}{2pt}
  \resizebox{\textwidth}{!}{%
  \begin{tabular}{l|cccc|ccc}
    \toprule
    & \multicolumn{4}{c|}{\textbf{H0 (Encoding)}} & \multicolumn{3}{c}{\textbf{H15 (Imagination)}} \\
    \cmidrule(lr){2-5} \cmidrule(lr){6-8}
    \textbf{Model} & Color & Shape & Size & Position MSE $\uparrow$& Color & Shape & Size \\
    \midrule
    DreamerV3 & \textbf{99.92} \std{0.01} & 69.37 \std{2.41} &  98.92 \std{0.93} &0.006 \std{0.001}&  \textbf{99.73} \std{0.11} &  28.95 \std{1.65} &  87.49 \std{14.10} \\
    \quad + dual-stream only & \textbf{100.0} \std{0.00} & 40.01 \std{1.13} & 51.55 \std{2.49} &\textbf{0.068} \std{0.003}& \textbf{99.98} \std{0.00} & 25.04 \std{0.10} & 33.31 \std{0.29} \\
    \quad + SS only & \textbf{99.89} \std{0.02} & 66.83 \std{1.67} & 99.39 \std{0.09} & 0.007 \std{0.000} & \textbf{99.70} \std{0.14} & 26.65 \std{0.53} & 92.73 \std{3.44} \\
    \quad + dual-stream + SS & \textbf{100.0} \std{0.01} & 43.65 \std{7.38} & 62.20 \std{9.26} & \textbf{0.059} \std{0.006} & \textbf{99.94} \std{0.03} & 25.27 \std{0.27} & 33.80 \std{0.54} \\
    \quad + attention only & \textbf{99.97} \std{0.01} & 92.95 \std{1.21} & \textbf{99.92} \std{0.02} &\textbf{0.060} \std{0.001}& \textbf{99.94} \std{0.01} & 56.22 \std{4.35} & \textbf{99.71} \std{0.13} \\
    \midrule
    \starwm{}  & \textbf{99.99} \std{0.00}& \textbf{99.92} \std{0.02}& \textbf{100.0}\std{0.00} &  \textbf{0.066} \std{0.004}&\textbf{100.0} \std{0.00}& \textbf{99.57} \std{0.18}& \textbf{99.99} \std{0.01}\\
    \midrule
    R2-Dreamer & 99.42 \std{0.35} & 37.92 \std{2.52} & 74.51 \std{2.92} & 0.028 \std{0.004} & 71.41 \std{8.11} & 25.38 \std{0.23} & 33.30 \std{0.14} \\
    \bottomrule
  \end{tabular}
  }
\end{table}

\subsection{Full results in DMC Environments}
\subsubsection{Per-task results}
\label{app:full_dmc}
Table~\ref{tab:dmc_full} reports full evaluation returns with standard deviations across 3 seeds for all DMC tasks and settings. For the Clean setting, we report two DreamerV3 reference values. DreamerV3 (Original Paper) lists the scores reported in \citet{hafner2025mastering}. DreamerV3 (PyTorch) lists the scores obtained by running the widely-used open-source PyTorch reimplementation under identical training budgets. \starwm{} is also built on top of this PyTorch codebase.

\begin{table}[h]
  \caption{\textbf{\ref{app:full_dmc}: DMC evaluation returns. (mean $\pm$ standard deviation over 3 seeds, at the end of training)} 
  VH: sequential video (3M steps). 
  VHR: random video (1M steps). 
  Best per (task, setting) in \textbf{bold}. }
  \label{tab:dmc_full}
  \centering
  \small
  \setlength{\tabcolsep}{3pt}
  \resizebox{\textwidth}{!}{%
  \begin{NiceTabular}{l| l| c c c c c c}
    \toprule
    & & Walker Walk& Walker Run & Cheetah Run & Cartpole & Finger Spin & Ball-in-cup \\

    \midrule
    \multirow{4}{*}{\textbf{Clean}} 
    &\makecell[l]{DreamerV3\\(Original Paper)} &  966 & 797 & 846 & 843 & 886 & 975\\
    &\makecell[l]{DreamerV3\\(Pytorch)} &954.40 \std{3.15} & 749.93 \std{10.64} & 894.54 \std{3.08}& 840.10 \std{31.89} & 554.74 \std{97.56}& 964.65 \std{1.53} \\
     &\starwm{}  & 944.06 \std{2.57}& 686.81 \std{39.45}&762.77 \std{54.33} & 863.66 \std{5.62}& 559.35 \std{92.24}&948.07 \std{9.50} \\
     
     \midrule
         \multirow{6}{*}{\textbf{VH}} 
    &DreamerV3 & 327.39 \std{68.25} & 95.36 \std{40.38} & 73.30 \std{8.46} & 160.13 \std{12.17} & 217.50 \std{119.63} & 118.26 \std{23.78} \\
    &DenoisedMDP & 93.14 \std{24.69} & 39.70 \std{2.91} & 60.93 \std{25.55} & 161.14 \std{7.17} & 53.96 \std{36.57} & 168.34 \std{46.43} \\
    &Iso-Dream++ &  98.16 \std{12.49}& 43.79 \std{6.89} &28.01 \std{14.61} &139.92 \std{5.77}  &  6.35 \std{0.71}&  15.76 \std{4.05}\\
     &R2-Dreamer & \textbf{936.49} \std{8.87} & 471.63 \std{39.33} & 426.69 \std{70.67} & 652.00 \std{82.78} & \textbf{600.12} \std{50.34} & \textbf{358.53} \std{405.18} \\
     &\starwm{}  & 904.66 \std{33.78} & 378.60 \std{190.65} & \textbf{452.77} \std{22.76} & 308.50 \std{10.44} & 464.76 \std{28.24} & 114.21 \std{27.46} \\
     &\starwm{}-R  & 856.98 \std{98.32} & \textbf{502.63} \std{146.12}  & 405.28 \std{53.30} & \textbf{770.81} \std{52.29} & 390.72 \std{25.92} & 303.74 \std{99.62} \\
         
         \midrule
         \multirow{6}{*}{\textbf{VHR}} 
    &DreamerV3 & 517.48 \std{68.05} & 191.37 \std{2.64} & 212.62 \std{45.31} & 143.89 \std{3.23} & 390.74 \std{21.53} & 23.12 \std{11.97} \\
    &DenoisedMDP & 183.85 \std{79.73} & 90.49 \std{20.30} & 54.64 \std{35.79} & 141.36 \std{18.50} & 44.08 \std{60.37} & 137.53 \std{52.80} \\
    &Iso-Dream++ &  209.89 \std{41.89}& 92.45 \std{24.10} & 56.02 \std{53.80} & 122.07 \std{7.67} & 95.86 \std{134.01}  & 52.18 \std{38.04} \\
     &R2-Dreamer & 158.39 \std{16.06} & 56.80 \std{4.20} & 150.12 \std{30.88} & 139.85 \std{6.57} & 97.99 \std{17.11} & 239.40 \std{34.30} \\
     &\starwm{}  & 851.24 \std{51.07} & 435.41 \std{73.46} & \textbf{468.81} \std{44.04} & 562.96 \std{50.04} & 414.15 \std{70.40} & 97.64 \std{35.66} \\
     &\starwm{}-R  & \textbf{872.66} \std{23.07} & \textbf{536.26} \std{30.91} & 432.32 \std{41.24} & \textbf{650.64} \std{102.31} & \textbf{459.80} \std{25.19} & \textbf{302.88} \std{33.24} \\
    \bottomrule
  \end{NiceTabular}
  }
\end{table}

\subsubsection{Aggregate Statistical Analysis}
\label{app:reliable_rl}

\begin{figure}[htbp]
    \centering
    
    \begin{subfigure}{\textwidth}
        \centering
        \includegraphics[width=0.8\linewidth]{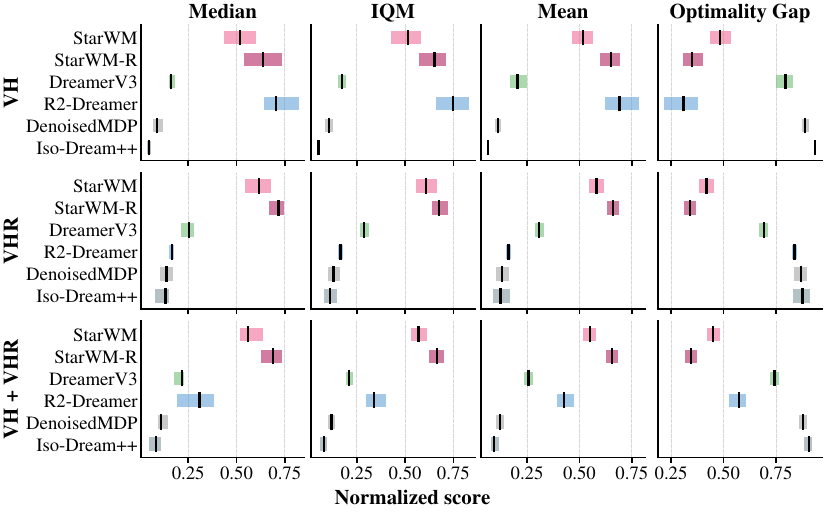}
        \caption{Aggregate point estimates with 95\% stratified bootstrap confidence intervals. Higher Median/IQM/Mean and lower Optimality Gap indicate stronger performance. Scores are normalized by DreamerV3 clean-environment performance.}
        \label{fig:all_aggregate}
    \end{subfigure}
    \vspace{1em}
    
    \begin{subfigure}{\textwidth}
        \centering
        \includegraphics[width=0.75\linewidth]{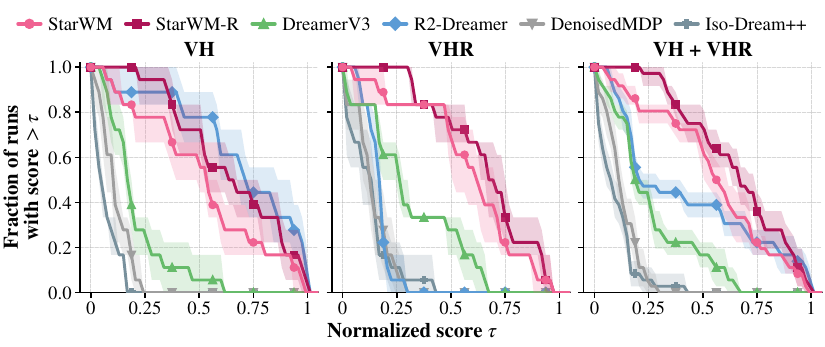}
        \caption{Performance profiles. Each curve plots the fraction of runs achieving normalized score above $\tau$; curves further to the upper right indicate stronger and more reliable performance. Shaded regions show 95\% stratified bootstrap confidence bands.}
        \label{fig:all_profile}
    \end{subfigure}
    \vspace{1em}
    
    \begin{subfigure}{\textwidth}
        \centering
        \includegraphics[width=0.8\linewidth]{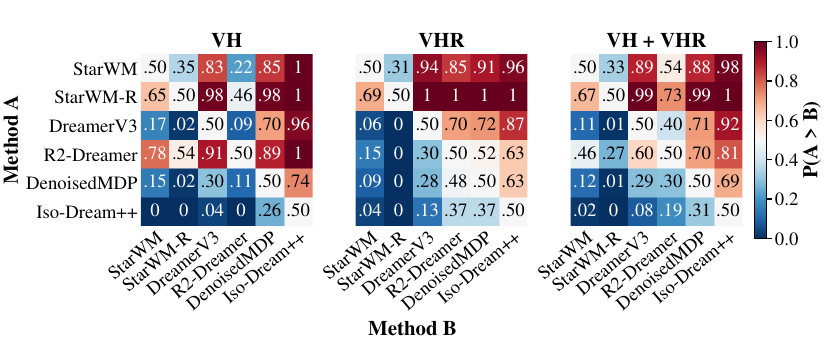}
        \caption{Pairwise probability of improvement $P(A > B)$, computed as the fraction of (run-of-$A$, run-of-$B$) pairs in which $A$ achieves a strictly higher score. Values above 0.5 indicate $A$ is more likely to outperform $B$ on a randomly drawn pair of runs.}
        \label{fig:all_poi_heatmap}
    \end{subfigure}
    
    \caption{\textbf{\ref{app:reliable_rl}: Aggregate statistical analysis on DMC under video distractors,} following~\citet{agarwal2021deep}. Each row reports VH, VHR, and the combined VH$+$VHR pool, computed over six DMC tasks and three random seeds (18 runs per method per setting; 36 in the pool).}
    \label{fig:reliable_rl_combined}
\end{figure}

To complement the per-task learning curves and tables in Appendices~\ref{app:full_dmc}, we report aggregate performance using the reliable RL evaluation framework of~\citet{agarwal2021deep}. All scores are normalized by DreamerV3's clean-environment performance on each task as a recovery ratio, then aggregated across the six DMC tasks and three random seeds (18 runs per method per setting). We report four point estimates, namely Median, Interquartile Mean (IQM), Mean, and Optimality Gap, each with 95\% stratified bootstrap confidence intervals (Figure~\ref{fig:all_aggregate}); performance profiles giving the fraction of runs exceeding each score threshold (Figure~\ref{fig:all_profile}); and pairwise probability-of-improvement heatmaps $P(A > B)$ (Figure~\ref{fig:all_poi_heatmap}). We provide each visualization separately for VH, VHR, and the combined VH$+$VHR pool.

\paragraph{Sequential video (VH).}
Under temporally coherent backgrounds, R2-Dreamer attains the highest aggregate scores across all four point estimates, with \starwm{}-R second and default \starwm{} third; the remaining baselines (DreamerV3, DenoisedMDP, Iso-Dream++) cluster at low IQM with large optimality gaps. The performance profile shows R2-Dreamer dominating \starwm{} across most score thresholds, while \starwm{}-R closes much of this gap. The pairwise heatmap reflects the same picture: $P(\text{\starwm{}-R} > \text{R2-Dreamer}) = 0.46$, statistically near the indifference threshold of 0.5, while $P(\text{\starwm{}} > \text{R2-Dreamer}) = 0.22$. As discussed in \S\ref{sec:exp_dmcgb}, the temporally coherent backgrounds in VH align well with R2-Dreamer's redundancy-reduction objective and partially co-opt \starwm{}'s contrastive guidance; reward prediction in \starwm{}-R substantially mitigates this gap.

\paragraph{Random video (VHR).}
Destroying the temporal coherence of the background reverses the ranking. \starwm{} and \starwm{}-R lead on every aggregate metric, while R2-Dreamer collapses to scores comparable to DreamerV3 and the latent-decomposition baselines. The performance profile shows both \starwm{} variants stochastically dominating all other methods across the full range of $\tau$. The pairwise heatmap confirms strong dominance: $P(\text{\starwm{}-R} > B) = 1$ for every non-\starwm{} baseline $B$, and $P(\text{\starwm{}} > B) \geq 0.85$ for every baseline except \starwm{}-R. R2-Dreamer's redundancy-reduction objective offers little robustness when the temporal structure it relies on is destroyed.

\paragraph{Cross-condition pool (VH$+$VHR).}
Pooling all 36 runs per method gives the cleanest measure of cross-condition robustness, since methods specialized to one distractor type are penalized for failing on the other. \starwm{}-R attains the highest aggregate score on all four point estimates and the lowest optimality gap, and stochastically dominates every baseline in the performance profile. \starwm{} is second on IQM, Mean, and Optimality Gap, behind only \starwm{}-R. R2-Dreamer's pool-level performance reflects its asymmetry: strong VH offset by weak VHR places it close to DreamerV3 in the IQM and Mean panels. The pairwise heatmap quantifies this: $P(\text{\starwm{}-R} > B) \geq 0.73$ for every non-\starwm{} baseline, and $P(\text{\starwm{}-R} > B) \geq 0.54$ except \starwm{}-R. Together, these pool-level results support the claim made in \S\ref{sec:exp_dmcgb} that \starwm{} and \starwm{}-R are the only methods robust across both distractor types.

\section{Ablation Studies \& Core Mechanistic Analysis}

\subsection{Reward as Additional Attention Routing Signal}
\label{app:reward_extension}

The default \starwm{} configuration uses inverse dynamics and contrastive learning to guide attention. Both signals target action-conditional or temporally predictable features. However, some task-relevant entities are neither directly action-conditional nor temporally distinctive: in \texttt{Ball-in-cup Catch}, the ball moves as a passive consequence of the cup's motion, and in \texttt{Cartpole Swingup}, the pole's angle is reward-critical but only indirectly controlled. Inverse dynamics provides weak gradients for such entities because their visual changes are not directly explained by the agent's action at a single timestep.

To address this, we augment the attention guidance with a reward prediction head that operates on the attention-pooled features $e_t^{\mathrm{attn}}$:
\begin{equation}
    \mathcal{L}_{\text{reward}}^{\text{attn}} = 
    \| \mathrm{MLP}_{\text{rwd}}(e_t^{\mathrm{attn}}) - r_t \|^2,
    \label{eq:reward_attn}
\end{equation}
with gradients flowing through the live attention map $A_t$ into the attention parameters, analogously to the inverse dynamics and contrastive losses. This directly incentivizes attention to cover regions whose visual content is informative of reward.

We report the results in the Figure~\ref{fig:dmc_curves} and the Table~\ref{tab:dmc_full}.

Reward prediction produces substantial improvements on tasks involving passive objects. On \texttt{Ball-in-cup Catch VHR}, the reward-augmented variant improves from 98 to 303, surpassing R2-Dreamer (239). On \texttt{Cartpole Swingup VH}, it improves from 308 to 771, also surpassing R2-Dreamer (652). In both cases, the passive object (ball, pole) is reward-critical but only indirectly action-conditional, and reward prediction provides the attention signal that inverse dynamics alone cannot.

These results demonstrate that \starwm{}'s architecture is agnostic to the specific attention loss: the gradient isolation mechanism ensures that any differentiable signal applied to the live attention path trains the routing without interfering with reconstruction. Reward prediction is one instance of a domain-specific signal that can be composed with or substituted for the default self-supervised objectives. The choice of attention loss thus becomes a modular design decision, selectable based on task characteristics: inverse dynamics for directly controlled entities, contrastive learning for temporally coherent entities, and reward prediction for passively task-relevant entities.

\subsubsection{Case Analysis: Attention Drift on \texttt{Cartpole Swingup VH}}
\label{app:failure_cartpole}
\begin{figure}[t]
    \centering
    \includegraphics[width=\textwidth]{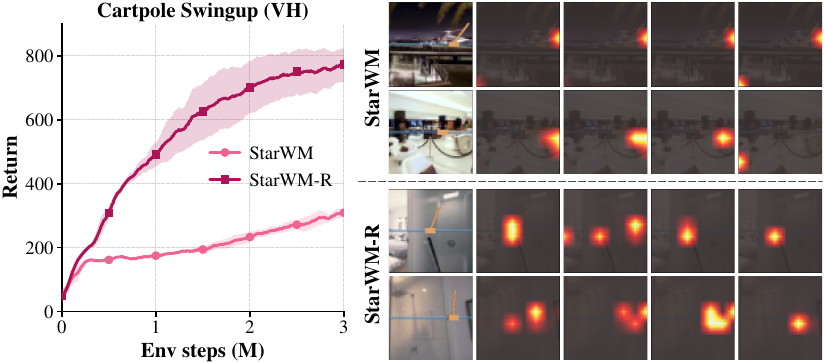}
    \caption{\textbf{\ref{app:failure_cartpole}: Default \starwm{}'s attention drifts on Cartpole VH; reward prediction recovers it.} Each row shows the input observation followed by the attention maps for the four queries. Default \starwm{} (top right) concentrates attention near the track endpoints without query specialization. \starwm{}-R (bottom right) localizes the cart and pole accurately, with queries specializing to distinct object regions.}
    \label{fig:cartpole_attention}
\end{figure}

\texttt{Cartpole Swingup VH} is the task where \starwm{} most strongly underperforms \starwm{}-R. Figure~\ref{fig:cartpole_attention} visualizes attention maps for default \starwm{} and \starwm{}-R on \texttt{Cartpole Swingup VH}. Default \starwm{}'s four queries concentrate near the left and right endpoints of the track rather than on the cart itself, and exhibit little specialization across queries. \starwm{}-R locates the cart and pole accurately, with queries specializing to distinct object regions. 

Several factors plausibly contribute to this failure: the small size of the cart-and-pole assembly relative to the $8 \times 8$ attention resolution, the temporal coherence of the sequential video background that competes with the cart for the contrastive objective's attention, and the lack of corrective signal from inverse dynamics in this configuration. When self-supervised proxies (controllability, temporal coherence) do not point unambiguously at the task-relevant object, attention can drift to other plausible regions. This pattern motivates the reward-augmented variant \starwm{}-R (Appendix~\ref{app:reward_extension}): when self-supervised signals underdetermine the routing target, reward provides a direct task-relevance signal that breaks the ambiguity. The same mechanism explains the \texttt{Ball-in-cup} improvement, where the passive ball is weakly supervised by inverse dynamics until reward fills the gap. Practitioners can consider adding reward prediction whenever the default configuration's attention maps appear to drift or fail to specialize on a particular task.

\subsection{Gradient Isolation Analysis}
\label{app:gradient_isolation}

The stop-gradient operator in \starwm{} serves two roles: blocking 
reconstruction gradients from the decoder pathway (\emph{detach~B}: 
$\bar{A}_t$ used in inverse broadcasting and spatial composition) and blocking 
RSSM gradients from the dynamics pathway (\emph{detach~A}: $\bar{A}_t$ used 
in computing entity tokens $e_t^{dyn}$); see Figure~\ref{fig:architecture} for where we apply detach. We evaluate configurations that remove these 
independently, with and without self-supervised attention guidance (SS), 
across environments of different distractor characteristics. The five configurations and the signals reaching the attention parameters in each are summarized in Table~\ref{tab:detach_configs}.

\begin{table}[ht]
\caption{\textbf{\ref{app:gradient_isolation}: Configurations evaluated in the gradient isolation analysis.} Each row indicates which gradient pathways reach the attention parameters. detach~A blocks RSSM gradients 
(via $\bar{A}_t$ in entity tokens); detach~B blocks reconstruction gradients (via $\bar{A}_t$ in the dual-stream decoder); SS denotes self-supervised attention guidance (inverse dynamics + contrastive). A check mark indicates the gradient or signal reaches the attention parameters.}
\label{tab:detach_configs}
\centering
\small
  \setlength{\tabcolsep}{5.5pt}
\begin{tabular}{clccc}
\toprule
\textbf{Symbol} &\textbf{Configuration} & \textbf{SS} & \textbf{RSSM grad.} & \textbf{Recon. grad.} \\
\midrule
\ding{72}&\starwm{} (full) & \checkmark & --- & --- \\
\ding{73}1&\quad no detach A/B, with SS & \checkmark & \checkmark & \checkmark \\
\ding{73}2&\quad no detach A/B, no SS & --- & \checkmark & \checkmark \\
\ding{73}3&\quad only detach B, no SS & --- & \checkmark & ---  \\
\ding{73}4&\quad only detach B, with SS & \checkmark & \checkmark & --- \\
\bottomrule
\end{tabular}
\end{table}

\paragraph{All configurations succeed on NoisyShape.}
\begin{figure}[t]
    \centering
    \includegraphics[width=\textwidth]{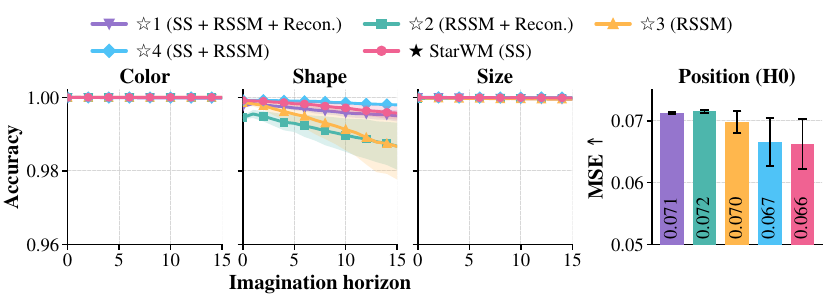}
\caption{\textbf{\ref{app:gradient_isolation}: Probe accuracy on NoisyShape under five gradient isolation configurations.} NoisyShape probe accuracy is invariant to gradient isolation choice.}
\label{fig:detach_noisyshape}
\end{figure}
We first evaluate all five configurations in Table~\ref{tab:detach_configs} on NoisyShape. Figure~\ref{fig:detach_noisyshape} shows that shape, color, and size retention at $H_{15}$ remain near optimal across all configurations: whether attention is trained by self-supervised signals alone, by reconstruction and RSSM gradients alone, or by any combination, the resulting attention reliably localizes the foreground shape. This invariance is consistent with the analysis in \S\ref{sec:exp_ablation}: in NoisyShape, the easiest object to reconstruct (a geometric shape on a stochastic background) through RSSM pathway coincides with the action-conditional and temporally predictable object, so reconstruction gradients and self-supervised signals push attention toward the same target. The architecture's structural capacity for spatial routing is sufficient on its own; gradient isolation does not become necessary until reconstruction gradients and self-supervised signals start to disagree.

\paragraph{Detach~B is critical under visual distractors.}
On DMC with video backgrounds, the alignment between reconstruction gradients and self-supervised signals breaks down: the visually dominant region (the video background) is no longer the same as the action-conditional region (the agent body). We evaluate three configurations from Table~\ref{tab:detach_configs} that test the necessity of each component: \ding{72}: full \starwm{} (SS as the sole attention signal), \ding{73}1: ``no detach A/B with SS'' (SS competing with both RSSM and reconstruction gradients on the attention path), and \ding{73}3: ``only detach B, no SS'' (RSSM gradient as the sole attention signal). Table~\ref{tab:dmc_detach} reports evaluation returns on Walker Walk and Cheetah Run under Clean and VHR settings.

\begin{table}[t]
\caption{\textbf{\ref{app:gradient_isolation}: Gradient isolation ablation on DMC.} Episode return 
(mean over 3 seeds) for Walker Walk and Cheetah Run under Clean and 
VHR settings. Configurations are defined in 
Table~\ref{tab:detach_configs}. Best per column in bold.}
\label{tab:dmc_detach}
\centering
\resizebox{\textwidth}{!}{%
\begin{tabular}{c l cc cc}
\toprule
& & \multicolumn{2}{c}{\textbf{Walker Walk}} 
& \multicolumn{2}{c}{\textbf{Cheetah Run}} \\
\cmidrule(lr){3-4} \cmidrule(lr){5-6} 
\textbf{Symbol}&\textbf{Configuration} & Clean & VHR & Clean & VHR \\
\midrule
\ding{72}&\starwm{} (full) & \textbf{952.37} \std{3.06} & \textbf{872.94} \std{50.99} & 752.37 \std{34.84} & \textbf{466.03} \std{53.00}  \\
\ding{73}1&no detach A/B, with SS & 810.66 \std{157.94} & 181.08 \std{17.72} & 641.66 \std{52.40} & 181.75 \std{31.91}  \\
\ding{73}3&only detach B, no SS & 943.42 \std{18.18} & 135.02 \std{20.74} & \textbf{880.96} \std{6.10} & 199.37 \std{94.74}  \\
\bottomrule
\end{tabular}}
\end{table}

\begin{figure}[t]
    \centering
    \includegraphics[width=\textwidth]{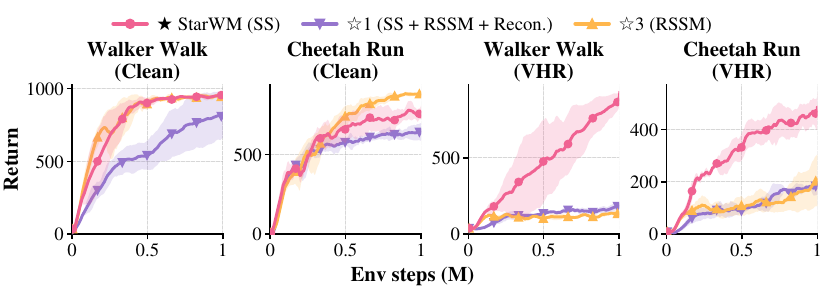}
    \caption{\textbf{\ref{app:gradient_isolation}: Learning curves for the gradient isolation ablation on DMC.} Episode return over training for the three configurations of Table~\ref{tab:dmc_detach}. Removing detach~B causes attention to be co-opted by reconstruction gradients; removing SS leaves RSSM gradients as the only attention signal, insufficient to disambiguate foreground from background under video distractors. Full \starwm{} combines detach~B with SS guidance.}
    \label{fig:detach_dmc}
\end{figure}

Removing detach~B (no detach A/B with SS; \ding{73}1) causes substantial degradation under VHR despite the presence of self-supervised signals. With reconstruction gradients reaching the attention parameters, the dense pixel loss pushes attention toward spatial coverage that minimizes reconstruction error, dragging attention toward the visually dominant background. SS gradients alone cannot counter this pull when reconstruction gradients are not blocked.

Removing SS while keeping detach~B (only detach B, no SS; \ding{73}3) also fails under VHR. Without explicit dynamics-grounded guidance, RSSM gradients are the sole attention signal; while RSSM does favor predictable features, this preference alone cannot reliably distinguish a controllable foreground from a temporally coherent background when both are reconstructable. Notably, this configuration performs comparably to or even better than full \starwm{} on Clean environments, where there is no background distractor to disambiguate, and adding SS introduces minor optimization overhead. Under distractors, however, the SS guidance becomes essential.

The two failure modes together establish that detach~B and SS are jointly necessary, not redundant: detach~B prevents reconstruction from co-opting the attention parameters, and SS provides the dynamics-grounded signal that correctly steers them. The cross-attention architecture provides the structural capacity for spatial routing, but the choice of what routes the attention requires both elements together when reconstruction and attention objectives compete.

\begin{figure}[t]
    \centering
    \includegraphics[width=0.75\textwidth]{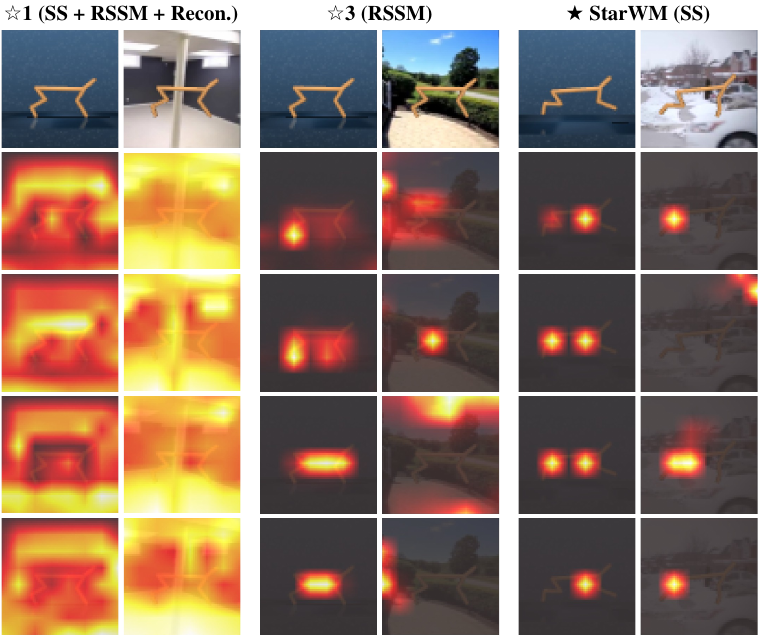}
    \caption{\textbf{\ref{app:gradient_isolation}: Gradient isolation determines attention focus on DMC.} Attention maps from the four learned queries on Walker Walk under VHR, comparing the three configurations of Table~\ref{tab:dmc_detach}. Top row: input observation. Rows 2--5: attention map for each of the four queries. Left (\ding{73}1: SS+RSSM+Recon., no detach): attention collapses to a nearly uniform distribution. Middle (\ding{73}3: RSSM only, detach B without SS): attention shows weak focus (1/4 queries specializing to agent body in VH). Right (\ding{72} \starwm{}: SS with detach A/B): attention is sharply localized on the agent's body parts (3/4 queries specializing to agent body in VH).}
    \label{fig:detach_dmc_attntion_map}
\end{figure}

\paragraph{Attention maps reveal the mechanism.}
To directly observe how each gradient pathway shapes attention behavior, we visualize the learned attention maps and quantify them via two statistics: the entropy of the attention distribution (lower = sharper) and the maximum attention weight (higher = more localized). Figure~\ref{fig:detach_dmc_attntion_map} shows attention maps for the four learned queries on Walker Walk, and Table~\ref{tab:attention_entropy} reports the corresponding quantitative statistics for $Q_0$ on Cheetah Run.

\begin{table}[t]
\caption{\textbf{\ref{app:gradient_isolation}: Attention map statistics under three gradient 
isolation configurations on Cheetah Run.} Average entropy and maximum attention weight of query $Q_0$, computed over evaluation trajectories. Lower entropy and higher max weight indicate sharper, more focused attention. The maximum possible entropy is $\log(64) \approx 4.16$ (uniform over the 64 spatial tokens), and the uniform floor for max attention is $1/64 \approx 0.016$.}
\label{tab:attention_entropy}
\centering
\begin{tabular}{cl cc cc}
\toprule
& & \multicolumn{2}{c}{\textbf{Clean}} 
& \multicolumn{2}{c}{\textbf{VHR}} \\
\cmidrule(lr){3-4} \cmidrule(lr){5-6} 
\textbf{Symbol}&\textbf{Configuration} & Entropy $\downarrow$ & Max $\uparrow$ & Entropy $\downarrow$ & Max $\uparrow$ \\
\midrule
\ding{72}&\starwm{} (full)        & \textbf{0.14} & \textbf{0.92} & \textbf{0.15} & \textbf{0.94} \\
\ding{73}1&no detach A/B, with SS  & 3.90          & 0.04          & 4.15          & 0.02          \\
\ding{73}3&only detach B, no SS    & 3.12          & 0.14          & 1.27          & 0.52          \\

\bottomrule
\end{tabular}
\end{table}

\ding{73}1 (no detach A/B with SS) produces nearly uniform attention. When all three gradient pathways (SS, RSSM, reconstruction) reach the attention parameters simultaneously, the dense pixel loss dominates: it pulls attention toward broad spatial coverage, defeating the routing mechanism entirely. Queries lose specialization, and the architecture's spatial bottleneck is bypassed.

\ding{73}3 (Only detach B, no SS) produces moderately structured attention. RSSM gradients alone favor temporally predictable regions, but this signal is too weak to specialize the four queries: under video distractors, only one of the four queries reliably localizes the agent, while the remaining three drift across various background regions. RSSM gradients identify some predictable region but cannot disambiguate among the multiple predictable regions present in a structured-video scene. On Clean Cheetah, though attention is even less focused, all of queries localize on the agent. This explains why this configuration's Clean Cheetah return exceeds \starwm{}. When there is nothing to filter, defocused attention loses no information and may benefit from broader feature coverage.

\emph{Full \starwm{}} produces sharply focused attention. The attention sharpness on Clean and VHR is nearly identical, indicating that SS guidance with stop-gradient isolation produces a routing decision that is independent of the visual complexity of the background. Equally important, three of the four queries reliably specialize to distinct agent body regions under video distractors, with only one query attending to the background. Compared to $1{:}3$ under RSSM-only guidance, this directly reflects the role of self-supervised signals. 

\subsection{Hyperparameter Sensitivity}
\label{app:n_sensitivity}
\label{app:hparam_sensitivity}

\paragraph{Default configuration.}
All hyperparameters remain fixed across every evaluated DMC task and distractor regime without per-task tuning. We set self-supervised loss weights to $\lambda_\mathrm{inv} = \lambda_\mathrm{ctr} = 1.0$ with temperature $\tau = 0.1$, and apply a fixed $\beta_\mathrm{bg}$ annealing schedule across all experiments. The sweeps reported below evaluate stability under post-hoc variations rather than serving as a parameter search.

\paragraph{Loss weights and KL schedule perturbations.}
We evaluate single-parameter perturbations on Walker Walk VHR at 1M steps across 3 seeds.

\begin{table}[h]

  \caption{\textbf{\ref{app:hparam_sensitivity}: One-at-a-time hyperparameter perturbations} (\starwm{}, Walker Walk VHR, matched 1M steps, mean $\pm$ SE, 3 seeds).}
  \label{tab:hparam_sensitivity}
  \centering
  \small
  \begin{tabular}{@{}l|c@{}}
    \toprule
    Perturbation & Return @ 1M \\
    \midrule
    $\lambda_\mathrm{inv} = 0.3$ ($3\times$ reduction)         & \textbf{925} \std{1} \\
    $\beta_\mathrm{bg}$: annealing removed entirely           & 918 \std{12} \\
    $\lambda_\mathrm{inv} = 3.0$ ($3\times$ increase)          & 856 \std{34} \\
    $\beta_\mathrm{bg}^{\mathrm{final}} = 2.0$ ($4\times$ increase) & 787 \std{77} \\
    \bottomrule
  \end{tabular}
\end{table}

Across all variations, performance remains solid between 787 and 925 (Table~\ref{tab:hparam_sensitivity}). Removing $\beta_\mathrm{bg}$ annealing entirely yields 918, showing that training does not depend strictly on schedule design. Higher penalties ($\lambda_\mathrm{inv} = 3.0$ and $\beta_\mathrm{bg}^{\mathrm{final}} = 2.0$) slow convergence speed within the 1M budget, but all configurations remain functional without diverging.

\subsubsection{Decomposition of Self-Supervised Objectives (Inv vs. Ctr)}
\label{app:ss_ablation}

\starwm{}'s default attention guidance combines inverse dynamics and contrastive learning. While Appendix~\ref{app:gradient_isolation} established that gradient isolation and self-supervised guidance are jointly necessary, this section asks a finer question: how do the two self-supervised signals divide the work? We characterize each signal's contribution on NoisyShape and across DMC, and identify a regime where contrastive learning introduces a partial conflict.

\begin{figure}[t]
    \centering
    \includegraphics[width=\textwidth]{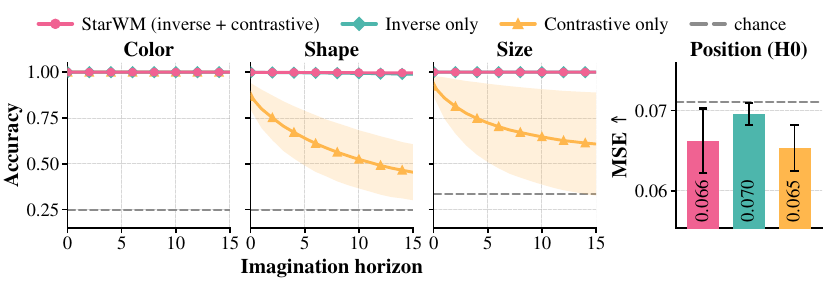}
    \caption{\textbf{\ref{app:ss_ablation}: Inverse dynamics is the primary attention 
    signal.} Probe accuracy on the RSSM latent for full 
    \starwm{} (inverse + contrastive), inverse-only, and 
    contrastive-only variants on NoisyShape. Inverse dynamics alone 
    is sufficient for near-perfect attribute retention; contrastive 
    alone fails to localize the foreground.}
    \label{fig:ss_ablation_noisyshape}
\end{figure}

On NoisyShape, we ablate the two attention guidance signals independently: inverse dynamics alone, contrastive alone, and the full combination. Figure~\ref{fig:ss_ablation_noisyshape} reports probe accuracy on the RSSM latent. Full \starwm{} and the inverse-dynamics-only variant achieve indistinguishable performance across all attribute probes (color, shape, size at $H_{15}$), while the contrastive-only variant fails to guide attention toward the foreground: shape and size retention degrades through imagination. This contrast reveals the qualitative difference between the two signals. Inverse dynamics provides direct supervision: the gradient points to regions whose visual changes are explained by the agent's action, an unambiguous target. Temporal contrastive learning provides only indirect supervision: it rewards features that are temporally consistent. In a randomized-background environment like NoisyShape, this indirect signal is too weak to specify the correct target. 

\begin{figure}[t]
    \centering
    \includegraphics[width=0.85\textwidth]{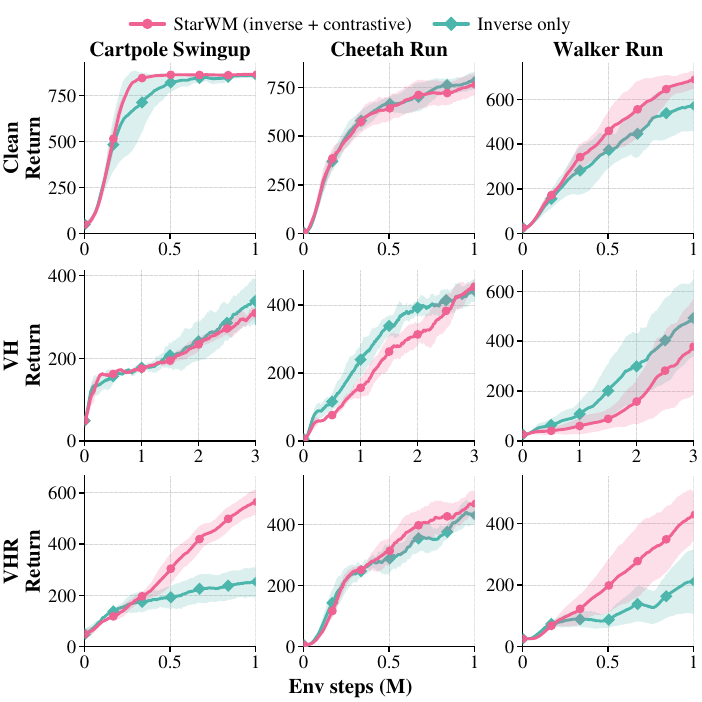}
    \caption{\textbf{\ref{app:ss_ablation}: Episode return for full \starwm{} (inverse + contrastive) versus inverse-only across Clean, VH (sequential video), and VHR (random video).} Contrastive supervision is essential under VHR, neutral under Clean and under VH.}
    \label{fig:ss_ablation_dmc}
\end{figure}

On DMC, we explore whether the contrastive signal is indeed helpful. Figure~\ref{fig:ss_ablation_dmc} compares full \starwm{} with the inverse-only variant across Clean, VH, and VHR settings.

Under VHR, contrastive learning is clearly beneficial: removing it substantially degrades performance. With independently sampled background frames at every step, the only feature with reliable temporal coherence is the agent itself. Contrastive learning exploits exactly this signal, providing dense temporal supervision in transitions where the agent's action is small or nearly action-independent (and thus where inverse dynamics gradient is weak). The two signals act complementarily: inverse dynamics identifies controllable regions, contrastive identifies temporally coherent regions, and under VHR these two sets coincide on the agent.

Under Clean and VH, the two configurations perform comparably.The inverse-only variant is even slightly better than full \starwm{} under VH. This is the regime where contrastive learning introduces a partial signal conflict: video frames played sequentially produce a temporally coherent background, satisfying the contrastive objective just as well as the agent does. Contrastive gradients then split ambiguously between agent and background, slightly diluting the otherwise correct routing established by inverse dynamics.

\paragraph{The fundamental tension: predictability without task 
information.}
The VH result reflects an epistemic limitation that is not specific to contrastive learning, but inherent to the self-supervised setting itself. When a background is temporally predictable, it is genuinely ambiguous whether the agent should encode it. In a task where reward depends on a moving landmark, the predictable background is task-relevant and should be retained; in a task where the agent must navigate around a distractor that happens to move predictably, it should be filtered. The agent has no way to tell these cases apart from self-supervised observation alone.

The weight of contrastive learning therefore encodes a prior held by the designer rather than a fact recoverable from the environment. A high weight reflects a prior that predictable structure should be preserved; a low weight reflects a prior that it should be filtered. This is a structural feature of self-supervision rather than a defect of the contrastive objective. Resolving it requires information beyond observation dynamics, such as a reward signal. We discuss in Appendix~\ref{app:reward_extension} how reward prediction can be added as an additional attention-guidance signal that breaks the ambiguity.

\subsubsection{Query count $N$.}
\begin{figure}[t]
    \centering
    \includegraphics[width=0.85\textwidth]{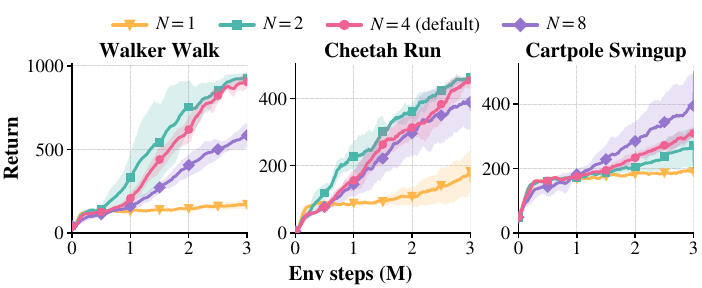}
    \caption{\textbf{\ref{app:n_sensitivity}: Return curves across varying query counts $N$ on DMC VH} (mean $\pm$ SE, 3 seeds).}
    \label{fig:ablation_N_curves}
\end{figure}

We evaluate the number of routing queries $N \in \{1, 2, 4, 8\}$ on three DMC tasks under Video Hard Sequential (VH) distractors at 3M steps (Table~\ref{tab:n_sensitivity} and Figure~\ref{fig:ablation_N_curves}).

\begin{table}[h]
  \caption{\textbf{\ref{app:n_sensitivity}: Effect of query count $N$ on DMC VH} (episode return at 3M steps, mean $\pm$ std, 3 seeds).}
  \label{tab:n_sensitivity}
  \centering
  \begin{tabular}{c|ccc}
    \toprule
    $N$ & Walker Walk & Cheetah Run & Cartpole Swingup \\
    \midrule
    1 & 170.48 \std{24.59} & 181.22 \std{63.81} & 191.54 \std{6.46} \\
    2 & \textbf{928.73} \std{34.39} & \textbf{461.71} \std{15.09} & 268.62 \std{60.34} \\
    4 & 902.31 \std{37.20} & 452.67 \std{21.67} & 308.00 \std{12.04} \\
    8 & 585.14 \std{72.84} & 387.97 \std{78.67} & \textbf{394.90} \std{98.07} \\
    \bottomrule
  \end{tabular}
\end{table}

A single query ($N{=}1$) underperforms across all tasks because a single attention mask cannot resolve distinct parts of an articulated body simultaneously. Increasing to $N{=}2$ and $N{=}4$ provides sufficient expressive capacity, achieving strong returns across tasks. Setting $N{=}8$ improves Cartpole Swingup but adds variance on Walker Walk. We use $N{=}4$ as a balanced default across all benchmarks.

\subsection{Selective Spatial Encoding Analysis}
\label{app:predictable_position}

\begin{figure}[t]
    \centering
    \includegraphics[width=0.8\textwidth]{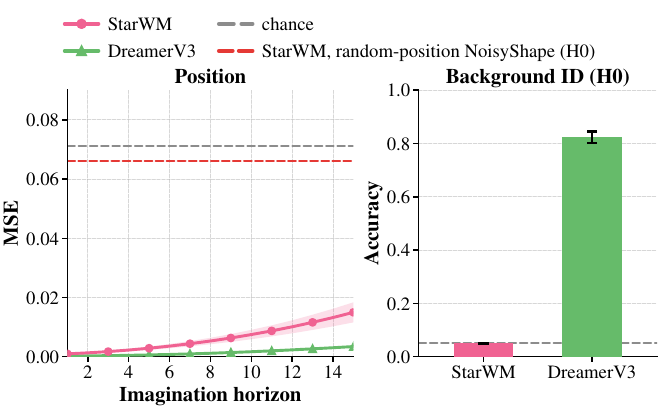}
\caption{\textbf{~\ref{app:predictable_position}: Latent space probing under predictable object motion.} Left: Position MSE across a 15-step imagination horizon. When position follows a deterministic trajectory, \starwm{} successfully tracks and rolls out spatial coordinates well below chance level. Right: Background identification accuracy at the encoding step (H0). DreamerV3 suffers from severe objective-irrelevant contamination, while \starwm{} perfectly drops background identity. The grey dashed line in the left panel is the chance level of the random-position NoisyShape (0.0711), shown for comparison with the red dashed line (\starwm{} on random-position NoisyShape); the chance level of this variant is 0.135 (Table~\ref{tab:predictable_position_restructured}).}

    \label{fig:predictable_position}
\end{figure}

The main NoisyShape experiments (Figure~\ref{fig:probe_curves} and Table~\ref{tab:noisyshape_full}) randomize position independently at each step, making it a purely non-predictable attribute that an ideal world model should discard. A natural concern is whether \starwm{}'s low position probe accuracy (MSE $=0.066$, near chance) reflects a correct identification of position as an irrelevant feature, or an architectural inability to encode spatial information at all. To disambiguate, we construct a variant in which position follows a deterministic trajectory and evaluate whether \starwm{}'s latent correctly encodes position in this setting. The results are shown in Figure~\ref{fig:predictable_position} and Table~\ref{tab:predictable_position_restructured}.

\paragraph{Probing protocol.}
We follow the same probe protocol as Appendix~\ref{app:probe_method}, extended with two additional probes: position (regression, MSE) and background ID (20-way classification, chance $=0.05$). All numbers are mean $\pm$ standard deviation across 3 seeds.

\begin{table}[t]
  \caption{\textbf{~\ref{app:predictable_position}: Probe accuracy under predictable position.} 
  Position follows uniform linear motion with boundary reflection. Background 
  ID is a 20-way classification over the 20 background frames of the pool. The position chance level is that of this variant (positions uniform on $[0.05, 0.95]^2$). 
  H0 = encoding; H15 = 15 steps of open-loop imagination.}
  \label{tab:predictable_position_restructured}
  \centering
  \small
  \setlength{\tabcolsep}{4pt}
  \begin{tabular}{l|l|ccccc}
    \toprule
    \textbf{Stage} & \textbf{Model} & \textbf{Color} & \textbf{Shape} & \textbf{Size} & \textbf{Position MSE} & \textbf{BG ID} $\downarrow$ \\
    \midrule
    \multirow{2}{*}{\textbf{H0 (Encoding)}} 
    & DreamerV3 & 100.00 \std{0.00} & 99.70 \std{0.03} & 99.96 \std{0.04} & 0.0003 \std{0.0002} & 82.44 \std{2.07} \\
    & \starwm{} & 100.00 \std{0.00} & 99.88 \std{0.06} & 99.99 \std{0.01} & 0.0008 \std{0.0002} & 5.02 \std{0.17} \\
    \midrule
    \multirow{2}{*}{\textbf{H15 (Imagination)}} 
    & DreamerV3 & 99.98 \std{0.03} & 97.46 \std{1.56} & 99.99 \std{0.00} & 0.0034 \std{0.0002} & --- \\
    & \starwm{} & 100.00 \std{0.00} & 99.71 \std{0.18} & 99.99 \std{0.01} & 0.0150 \std{0.0046} & --- \\
    \midrule
    \multicolumn{2}{c|}{Chance level} & 25.00 & 25.00 & 33.33 & 0.135 & 5.00\\
    \bottomrule
  \end{tabular}
\end{table}

\paragraph{\starwm{} encodes predictable position.}
In contrast to the random-position setting where \starwm{} retains position only coarsely (MSE $\approx 0.066$, near chance), under predictable motion \starwm{}'s position MSE drops by two orders of magnitude to $\sim 0.0008$ at encoding and remains well below chance after 15 imagination steps. This confirms that the architecture does not blindly discard spatial information: when position is dynamically predictable, it is preserved by the RSSM latent. The slight gap to DreamerV3 at H15 reflects a reasonable trade-off, as \starwm{} routes spatial information through a bottleneck of $N=4$ attention queries rather than a dense feature map, introducing mild quantization at the price of filtering objective-irrelevant content.

\paragraph{Background ID separates the two methods cleanly.}
The background ID probe provides a direct readout of objective-irrelevant contamination. DreamerV3's latent classifies the background with $82.4\%$ accuracy over 20 classes, confirming that its dense reconstruction objective forces the encoder to preserve background identity even when the background carries no predictive information. \starwm{}'s latent achieves $5.0\%$ (exactly chance level), demonstrating that the dual-stream architecture successfully routes background content to the auxiliary VAE stream. This holds regardless of position predictability, indicating that \starwm{}'s filtering is driven by the architectural routing rather than by the motion pattern of the specific entity.

\paragraph{Summary.}
The two NoisyShape variants together establish the correct dual behavior: \starwm{} discards position when it is irrelevant and encodes position when it is relevant. DreamerV3 encodes position in both settings, and encodes background identity in both settings.

\subsection{Progressive Distraction Analysis across NoisyShape Variants}
\label{app:progressive_distraction}

\begin{figure}[t]
    \centering
    \includegraphics[width=0.8\textwidth]{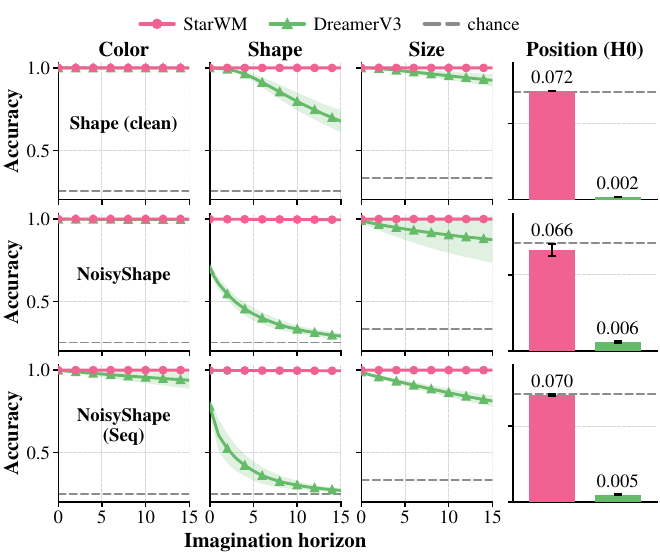}
\caption{\textbf{\ref{app:progressive_distraction}: Latent space probing across NoisyShape variants.} DreamerV3's performance degrades as irrelevant complexity increases, while \starwm{} is invariant to distraction complexity.}
\label{fig:paper_probe_comparison_all}
\end{figure}

To understand how irrelevant visual complexity affects latent representations, we compare probe accuracy across NoisyShape variants of increasing difficulty, plus a distraction-free baseline. The results are shown in Figure~\ref{fig:paper_probe_comparison_all} and Table~\ref{tab:progressive_distraction_stacked}.

\begin{table}[h]
  \caption{\textbf{\ref{app:progressive_distraction}: Probe accuracy across NoisyShape variants.} 
  H0 = encoding step; H15 = 15-step open-loop imagination. 
  All values: mean $\pm$ std over 3 seeds. Accuracies in \%, position in MSE.}
  \label{tab:progressive_distraction_stacked}
  \centering
  \begin{tabular}{ll|cccc}
    \toprule
    \textbf{Environment} & \textbf{Model} & \textbf{Color} & \textbf{Shape} & \textbf{Size} & \textbf{Pos. MSE $\uparrow$} \\
    \midrule
     \multicolumn{6}{l}{\textbf{\textit{H0 (Encoding)}}} \\
     \midrule
    Shape (fixed) & DreamerV3 & 100.00 & 100.00 & 100.00 & --- \\
    \midrule
    \multirow{2}{*}{Shape (clean)} 
    & DreamerV3 & 99.99 \std{0.01} & 99.87 \std{0.05} & 99.96 \std{0.01} & 0.002 \\
    & \starwm{} & 100.00 \std{0.00} & 100.00 \std{0.00} & 100.00 \std{0.00} & 0.072 \\
    \midrule
    \multirow{2}{*}{NoisyShape} 
    & DreamerV3 & 99.92 \std{0.01} & 69.37 \std{2.41} & 98.92 \std{0.93} & 0.006 \\
    & \starwm{} & 99.99 \std{0.00} & 99.92 \std{0.02} & 100.00 \std{0.00} & 0.066 \\
    \midrule
    \multirow{2}{*}{\makecell[l]{NoisyShape\\(Seq)}} 
    & DreamerV3 & 99.88 \std{0.05} & 78.02 \std{5.63} & 98.94 \std{0.18} & 0.005 \\
    & \starwm{} & 100.00 \std{0.00} & 99.82 \std{0.04} & 99.99 \std{0.00} & 0.070 \\
    \midrule
    \multicolumn{2}{c|}{Chance level} & 25.00 & 25.00 & 33.33 & 0.071 \\
    
    \midrule
    \midrule
     \multicolumn{6}{l}{\textbf{\textit{H15 (Imagination)}}} \\
     \midrule
    Shape (fixed) & DreamerV3 & 100.00 & 100.00 & 100.00 & --- \\
    \midrule
    \multirow{2}{*}{Shape (clean)} 
    & DreamerV3 & 99.93 \std{0.06} & 67.63 \std{6.60} & 92.48 \std{3.79} & --- \\
    & \starwm{} & 100.00 \std{0.00} & 99.98 \std{0.01} & 99.99 \std{0.01} & --- \\
    \midrule
    \multirow{2}{*}{NoisyShape} 
    & DreamerV3 & 99.73 \std{0.11} & 28.95 \std{1.65} & 87.49 \std{14.10} & --- \\
    & \starwm{} & 100.00 \std{0.00} & 99.57 \std{0.18} & 99.99 \std{0.01} & --- \\
    \midrule
    \multirow{2}{*}{\makecell[l]{NoisyShape\\(Seq)}} 
    & DreamerV3 & 93.88 \std{5.11} & 27.00 \std{0.87} & 81.25 \std{3.36} & --- \\
    & \starwm{} & 100.00 \std{0.00} & 99.63 \std{0.15} & 99.99 \std{0.00} & --- \\
    \midrule
    \multicolumn{2}{c|}{Chance level} & 25.00 & 25.00 & 33.33 & --- \\
    \bottomrule
  \end{tabular}
\end{table}

\paragraph{DreamerV3 degrades progressively with distraction complexity.}
The Shape (fixed) baseline confirms that DreamerV3's architecture is fully capable of encoding objective-relevant attributes: with no position variation and no background, all probes achieve 100\% accuracy through H15. However, performance degrades as irrelevant complexity increases. In Shape (clean), where the only irrelevant factor is random position, shape H15 already drops to 67.6\%. Adding random video backgrounds (NoisyShape) further reduces shape H0 from 99.9\% to 69.4\% and H15 to 29.0\%. This progressive collapse demonstrates that DreamerV3's dense reconstruction objective forces the RSSM to allocate representational capacity to objective-irrelevant content, and this allocation scales with the visual complexity and temporal structure of the distraction.

\paragraph{\starwm{} is invariant to distraction complexity.}
Across all three distraction levels, \starwm{} maintains near-perfect probe accuracy: shape H15 $\geq 99.6\%$, size H15 $\geq 99.99\%$, and color H15 $= 100.0\%$. The consistency from Shape (clean) through NoisyShape (Seq) confirms that the dual-stream architecture with attention-based routing effectively isolates objective-relevant from objective-irrelevant content regardless of the distraction's visual complexity or temporal structure.

\paragraph{Position MSE confirms objective-irrelevant filtering.}
In all variants with random position, \starwm{}'s position MSE remains near the chance level, while DreamerV3 encodes position with high precision.

\section{Stress Testing, Generalization \& Boundary Cases}

\subsection{Spatial Granularity \& Small Objects}
\label{app:small_objects}

Our default attention resolution of $8{\times}8$ spatial tokens (Appendix~\ref{app:algorithm}) balances spatial granularity with the computational cost of cross-attention. As flagged in \S\ref{sec:conclusion}, this resolution may be insufficient when task-relevant objects occupy very few pixels. To test this directly, we rescale Cheetah Run and Walker Walk so that the agent's body occupies approximately one quarter of the pixel area in the default rendering (i.e., linear half-size), and re-evaluate \starwm{}, DreamerV3, and R2-Dreamer under Clean, VH, and VHR settings.

\begin{figure}[t]
    \centering

    \begin{minipage}{0.69\textwidth}
        \centering
        \includegraphics[width=\textwidth]{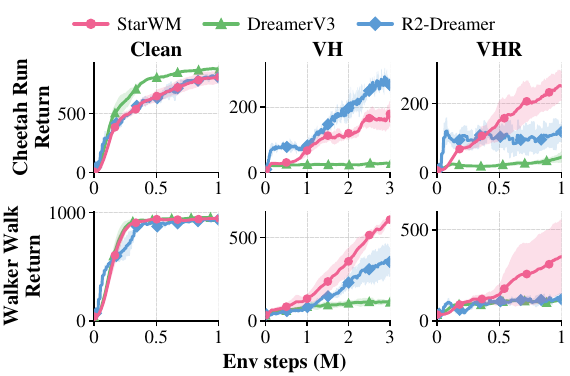}
    \end{minipage}
    \hfill 
    \begin{minipage}{0.29\textwidth}
        \centering
        \includegraphics[width=\textwidth]{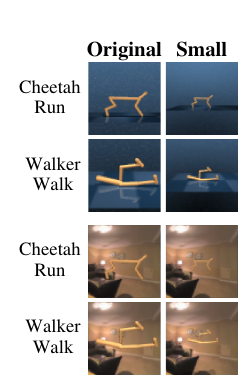} 
    \end{minipage}

    \caption{\textbf{\ref{app:small_objects}: Small-object stress test.} \textbf{\textit{Left:}} Learning curves for Cheetah Run and Walker Walk with the agent rescaled to half its default linear size (one quarter pixel area), under Clean, VH (3M steps), and VHR (1M steps) settings. \textbf{\textit{Right:}} Illustration of the environment setup.}
    \label{fig:small_objects}
\end{figure}

In clean environments, \starwm{} and DreamerV3 reach returns comparable to the original-scale clean setting, indicating that smaller object size alone does not prevent learning. Under distractors, all three methods degrade relative to their original-scale numbers; for example, \starwm{}'s VHR Walker Walk return drops from approximately 850 to 350. The shared degradation across methods suggests a regime where the visual signal-to-distractor ratio approaches the limits of these architectures, rather than a \starwm{}-specific failure mode.

The relative ordering between methods is largely preserved. Under VHR, \starwm{} continues to lead on both tasks despite the reduced absolute returns. Under VH, the result is split: \starwm{} dominates on Walker Walk while R2-Dreamer leads on Cheetah Run. Based on the original-scale results in \S\ref{sec:exp_dmcgb}, we expect reward-augmented attention to mitigate the small-object VH degradation, but verification remains for future work.

These results empirically confirm the limitation discussed in \S\ref{sec:conclusion}: spatial granularity matters when objects shrink below what $8\times8$ attention can localize cleanly. A natural architectural extension is to intercept the encoder one stage earlier to expose $16\times16$ spatial tokens, trading compute for finer routing. We expect this together with reward-augmented attention to partially recover the small-object regime, with the relative contribution of each remaining an open empirical question.

\subsection{Multi-Object Extension: NoisyTwoShape}
\label{app:multi_object}

We extend NoisyShape to a two-object setting to test (i) whether \starwm{}'s attention queries spontaneously partition across multiple foreground entities without architectural modification, and (ii) whether the resulting representation cleanly retains both objects through long-horizon imagination under visual distractors. The NoisyTwoShape environment renders two non-overlapping geometric shapes, one large and one small, on DAVIS video backgrounds re-sampled each step, with positions also re-sampled each step. Each shape has its own shape and color attributes controlled by an 8-action discrete space. We use the same \starwm{} configuration as in NoisyShape (4 attention queries, no hyperparameter tuning) and probe color, shape, and position for both objects (probe sets of 200 episodes for H0 and 100 episodes for imagination).

\begin{figure}[t]
    \centering
    \includegraphics[width=\linewidth]{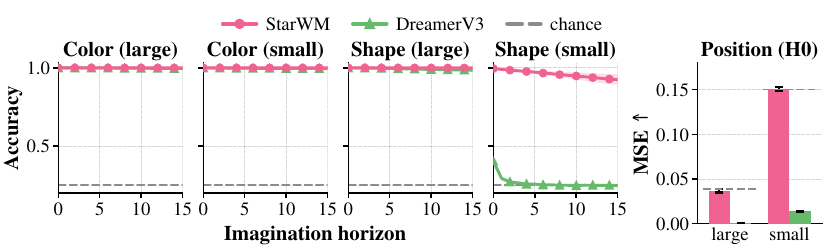}
    \caption{\textbf{\ref{app:multi_object}: Latent probe results on
    NoisyTwoShape.} Probe accuracy for color, shape 
    of the large and small object across H0 to H15 imagination horizon, 
    and position MSE at H0. Higher MSE on position indicates correct 
    irrelevant treatment (positions are re-sampled each step). Dashed lines indicate per-object chance levels for position MSE, computed via Monte Carlo simulation over each object's position distribution under its size-dependent margin: 0.038 (large) and 0.150 (small).}
    \label{fig:twoshape_probe}
\end{figure}

\begin{figure}[t]
    \centering
    \includegraphics[width=0.5\linewidth]{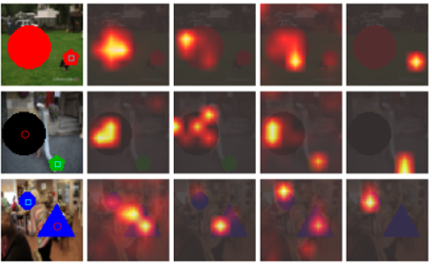}
    \caption{\textbf{\ref{app:multi_object}: \starwm{} attention maps on NoisyTwoShape.} Columns 
    show the input image followed by the four query attention maps. }
    \label{fig:twoshape_attention}
\end{figure}

\paragraph{Queries partition across objects.}
Figure~\ref{fig:twoshape_attention} shows that \starwm{}'s four attention queries distribute across the two foreground objects without per-object supervision, extending the emergent specialization observed in single-object NoisyShape to multi-object scenes. Across samples, queries reliably cover both the larger and the smaller object, and none drift to the video background despite the changing distractor.

\paragraph{Both objects retained through long-horizon imagination.}
Figure~\ref{fig:twoshape_probe} shows that this spatial partitioning translates into attribute retention through imagination. \starwm{} preserves color and shape of both objects with $\geq$99\% accuracy at the encoding step (H0), and retains them through 15-step open-loop imagination: both colors and the larger object's shape remain near 100\%, while the smaller object's shape degrades only modestly (from $\sim$99\% to $\sim$92\%). DreamerV3 by contrast fails on the smaller object from the encoding step itself: its small-shape probe reaches only 40\% at encoding (chance 25\%) and falls to chance during imagination, indicating that the encoder, which is forced by reconstruction to allocate capacity by pixel area, does not retain a usable representation of the smaller object even at the moment of observation. The reconstruction bias, mild for the larger object whose pixel signature dominates, becomes severe for the smaller one.
\paragraph{Position is correctly treated as objective irrelevant.}
The position MSE at H0 confirms that \starwm{} treats both object positions as objective-irrelevant, with MSE matching the chance reference for both the larger and the smaller object (chance levels differ across objects due to size-dependent position sampling ranges; computed by Monte Carlo simulation, see Figure~\ref{fig:twoshape_probe} caption). DreamerV3 by contrast encodes both positions with high precision, reflecting the dense reconstruction objective's tendency to retain spatially-localized identifiers regardless of their predictive value.

\subsection{Texture, Low Contrast, and Photorealistic Scenes}
\label{app:separability}

The main paper evaluates setups where the foreground visually stands out from the video background. Here, we test whether \starwm{}'s routing relies on this visual contrast or truly learns from the underlying dynamics. We evaluate three settings with reduced foreground--background contrast:
(i) a texture ladder that matches foreground and background visual distributions while keeping ground-truth labels;
(ii) a desaturated Walker agent on DMC; and
(iii) CALVIN, a photorealistic manipulation dataset where the robot, objects, table, and kitchen share the same visual render.

\paragraph{(i) Textured foreground ladder.}
We replace the solid color of the NoisyShape object with textures of increasing difficulty:
\textbf{L1} adds procedural textures from an external domain;
\textbf{L2} samples textures directly from the DAVIS background videos, sharing the same visual domain;
\textbf{L3} alpha-blends the L2 texture into the background ($\alpha=0.5$), removing clear boundaries.
All other factors remain unchanged, allowing direct use of our probe protocol (Appendix~\ref{app:probe_method}).

\begin{figure}[h]
    \centering

    \includegraphics[width=0.72\linewidth]{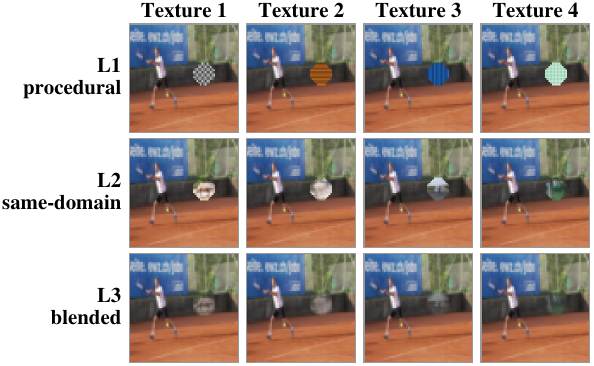}
    \caption{\textbf{\ref{app:separability}: Foreground texture ladder.} The same underlying state rendered across levels for the four texture IDs (replacing color). Only foreground appearance changes: L1 uses external procedural textures, L2 uses crops from DAVIS background videos, and L3 alpha-blends these crops into the background ($\alpha{=}0.5$, feathered edges).}
    \label{fig:textured_ladder_envs}
\end{figure}

\begin{figure}[h]
    \centering

    \includegraphics[width=\linewidth]{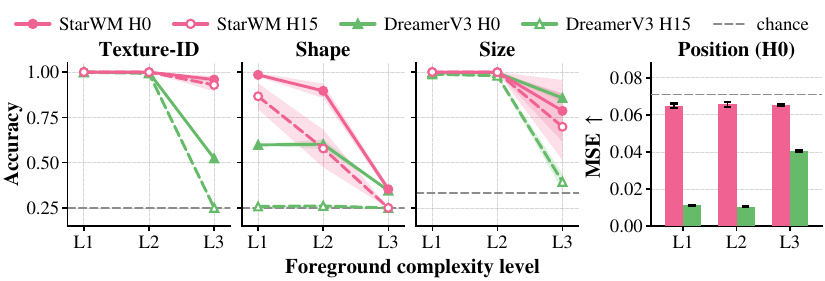}
    \caption{\textbf{\ref{app:separability}: Latent probe accuracy across the ladder} (mean $\pm$ SE, 3 seeds, 400k steps). Solid lines/filled markers denote the encoding step (H0); dashed lines/open markers denote 15-step open-loop rollouts (H15). Grey dashed lines show chance performance. For position MSE (right panel), higher values indicate that task-irrelevant position details are successfully filtered out. \starwm{} stays at chance, whereas DreamerV3 encodes exact position.}
    \label{fig:textured_probe_ladder}
\end{figure}

\begin{table}[h]

  \caption{\textbf{\ref{app:separability}: Probe accuracy on the texture ladder} (3 seeds, 400k steps). H0 = encoding step, H15 = 15-step rollout. Chance is 0.25 for texture and shape, 0.33 for size; chance position MSE is $\approx$0.071.}
  \label{tab:textured_ladder}
  \centering
  \small
  \begin{tabular}{ll|cc|cc|cc|c}
    \toprule
    & & \multicolumn{2}{c|}{texture-id} & \multicolumn{2}{c|}{shape} & \multicolumn{2}{c|}{size} & pos.\ MSE \\
    Level & Model & H0 & H15 & H0 & H15 & H0 & H15 & H0 \\
    \midrule
    \multirow{2}{*}{L1} & \starwm{}   & \textbf{1.000} & \textbf{1.000} & \textbf{0.984} & \textbf{0.866} & 1.000 & 1.000 & 0.065 \\
                        & DreamerV3   & 0.999 & 0.999 & 0.598 & 0.259 & 0.996 & 0.987 & 0.011 \\
    \midrule
    \multirow{2}{*}{L2} & \starwm{}   & \textbf{0.999} & \textbf{0.999} & \textbf{0.896} & \textbf{0.579} & 0.999 & 0.999 & 0.067 \\
                        & DreamerV3   & 0.997 & 0.993 & 0.600 & 0.262 & 0.996 & 0.980 & 0.010 \\
    \midrule
    \multirow{2}{*}{L3} & \starwm{}   & \textbf{0.959} & \textbf{0.929} & 0.354 & 0.251 & \textbf{0.786} & \textbf{0.698} & 0.065 \\
                        & DreamerV3   & 0.523 & 0.252 & 0.347 & 0.250 & 0.857 & 0.395 & 0.041 \\
    \bottomrule
  \end{tabular}
\end{table}

As shown in Figure~\ref{fig:textured_ladder_envs}, Figure~\ref{fig:textured_probe_ladder}, and Table~\ref{tab:textured_ladder}, both models maintain texture identity at L1 and L2 because the object still has defined boundaries. However, differences emerge on \emph{shape}: \starwm{} retains shape features over 15 imagined steps (0.866 at L1, 0.579 at L2), whereas DreamerV3 drops to chance ($\approx$0.25). 

At \textbf{L3}, where the boundary is blended, the difference in texture tracking is pronounced: DreamerV3 falls to chance (0.252) at H15, while \starwm{} maintains 0.929. Blending removes sharp contours, so shape accuracy drops for both models. Across all levels, \starwm{} retains object size better (0.698 vs.\ 0.395 at L3, H15) and keeps position MSE near chance (0.065 vs.\ 0.071), confirming that it ignores task-irrelevant spatial information. In contrast, DreamerV3 tracks position closely (0.041).

\emph{Attention routing vs.\ representation.} The attention mass on the object ($\approx$12\% of total pixels) shifts from 0.82 to 0.57 and 0.37 across L1--L3 (7.0$\times$, 4.9$\times$, and 3.2$\times$ above uniform attention). Because entity tokens use soft spatial weighting, \starwm{} tracks foreground identity reliably even when visual boundaries blur.

\paragraph{(ii) Low-contrast agent on DMC.}
We desaturate the Walker body colors to match the video background tone, evaluating performance across both distractor settings over 3 seeds.

\begin{table}[h]

  \caption{\textbf{\ref{app:separability}: Low-contrast Walker Walk return} (mean $\pm$ SE, 3 seeds). VH = coherent video background, VHR = per-step randomized frames.}
  \label{tab:lowcontrast}
  \centering
  \begin{tabular}{l|ccc}
    \toprule
    Setting & \starwm{} & DreamerV3 & R2-Dreamer \\
    \midrule
    Walker VH  \,(3M steps) & 893 \std{41} & 198 \std{22} & 895 \std{41} \\
    Walker VHR (1M steps)   & \textbf{890} \std{17} & 394 \std{34} & 118 \std{13} \\
    \bottomrule
  \end{tabular}
\end{table}

\begin{figure}[h]
    \centering
    \includegraphics[width=0.7\linewidth]{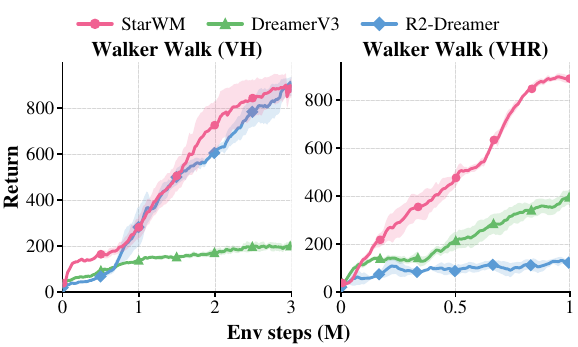}
    \caption{\textbf{\ref{app:separability}: Return curves for low-contrast Walker Walk} (mean $\pm$ SE, 3 seeds).}
    \label{fig:lowcontrast_walker}
\end{figure}

Under random frame distractors (VHR), \starwm{} reaches 890, substantially outperforming DreamerV3 (394) and R2-Dreamer (118). Under coherent video backgrounds (VH), \starwm{} (893) matches R2-Dreamer (895), while DreamerV3 reaches only 198. Removing color contrast does not impair \starwm{}'s robustness under rapid visual shifts.

\paragraph{(iii) Photorealistic manipulation (CALVIN).}
We evaluate \starwm{} on the CALVIN benchmark (\texttt{task\_D\_D}, 512k frames of teleoperated play, static camera, 7-DOF actions) for 200k offline steps. Here, the robot, table, blocks, and environment are part of a unified photorealistic render. We project the gripper and block centers into image space using camera calibration and measure the attention allocated to each region. To account for workspace framing, we evaluate each region against a shuffled null baseline (the same attention map applied to a different frame).

\begin{table}[h]

  \caption{\textbf{\ref{app:separability}: Spatial attention concentration on CALVIN.} Measured on frames where a block moves ($n=204$). Active blocks are identified by ground-truth movement.}
  \label{tab:calvin_attention}
  \centering
  \small
  \begin{tabular}{l|cccc}
    \toprule
    Region & Frame share & Attention share & Concentration & Shuffled null \\
    \midrule
    Gripper tip ($r=5$px)             & 1.9\% & 16.6\% & \textbf{8.69$\times$} & 2.83$\times$ \\
    Active block                      & 0.7\% & 4.2\%  & \textbf{6.09$\times$} & 2.56$\times$ \\
    Static blocks                     & 1.4\% & 2.1\%  & 1.52$\times$          & 2.03$\times$ \\
    Gripper + all blocks              & 3.3\% & 18.8\% & 5.63$\times$          & 2.48$\times$ \\
    \bottomrule
  \end{tabular}
\end{table}

\begin{figure}[h]
    \centering
    \includegraphics[width=0.7\linewidth]{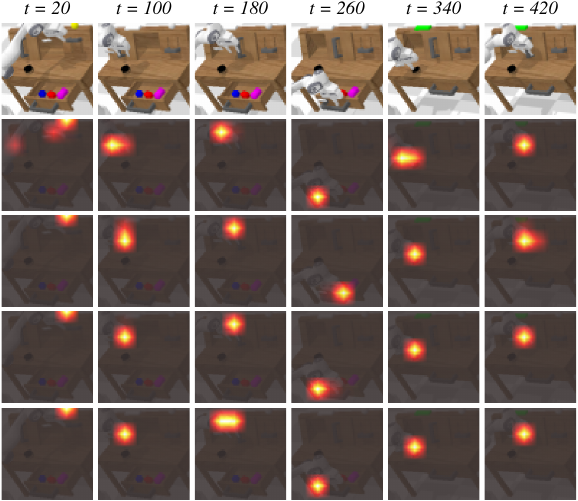}
    \caption{\textbf{\ref{app:separability}: Attention routing on CALVIN.} Input frame and corresponding attention maps from four queries. Attention targets the controllable objects and gripper over background structures.}
    \label{fig:calvin_attention}
\end{figure}

Attention focuses heavily on the gripper and the active block (8.69$\times$ and 6.09$\times$ concentration, compared to 2.83$\times$ and 2.56$\times$ nulls). In contrast, static blocks remain at chance levels (1.52$\times$ vs.\ 2.03$\times$ null), reflecting only spatial priors. Across all validation frames, gripper concentration averages 9.71$\times$ (2.84$\times$ null). Even without visual separation between foreground and background, \starwm{} routes attention based on controllability.

\subsection{Passive and Non-Action-Driven Task-Relevant Objects}
\label{app:passive}

\starwm{} trains its router using inverse dynamics and a temporal contrastive loss. Because both objectives favor directly controlled entities, we investigate whether the model retains task-relevant objects that are \emph{passive} (i.e., unaffected by agent actions). We evaluate two cases: a diagnostic environment with an explicitly static object, and DMC Reacher, where the target is a fixed goal rather than an actuated joint.

\begin{figure}[h]
    \centering
    \includegraphics[width=\linewidth]{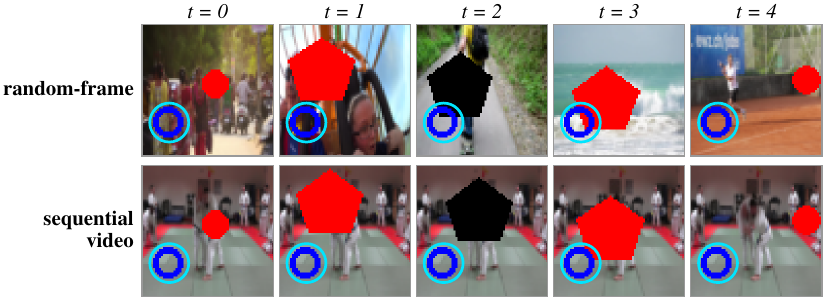}
    \caption{\textbf{\ref{app:passive}: \texttt{PassiveShapeEnv}.} Five consecutive steps under each background setting. The active object $A$ changes attributes based on actions and moves at each step; the passive object $B$ (cyan circle) maintains fixed attributes throughout the episode. Top: per-step randomized background frames. Bottom: sequential video background.}
    \label{fig:passive_env}
\end{figure}

\paragraph{(i) Diagnostic environment: \texttt{PassiveShapeEnv}.}
We augment NoisyShape with a second object $B$ alongside the controllable object $A$. Object $B$'s color and position are sampled once per episode and remain fixed regardless of agent actions (Figure~\ref{fig:passive_env}). Because actions never alter $B$, it provides no inverse-dynamics gradient. We evaluate whether $B$'s properties are preserved in the latent state using the probing protocol in Appendix~\ref{app:probe_method}, evaluating continuous position via regression ($R^2$).

\begin{table}[h]
  \caption{\textbf{\ref{app:passive}: Latent probe recovery for passive object $B$} (mean $\pm$ SE, 3 seeds, 212k steps). $b$\_colour is 4-way classification accuracy (chance = 0.25); $b$\_position reports held-out $R^2$.}
  \label{tab:passive}
  \centering
  \small
  \begin{tabular}{l|cc|cc|cc}
    \toprule
    & \multicolumn{2}{c|}{$b$\_colour (acc)} & \multicolumn{2}{c|}{$b$\_$x$ ($R^2$)} & \multicolumn{2}{c}{$b$\_$y$ ($R^2$)} \\
    Background & H0 & H15 & H0 & H15 & H0 & H15 \\
    \midrule
    random-frame      & \textbf{1.000} \std{.000} & \textbf{1.000} \std{.000} & \textbf{0.80} \std{.09} & \textbf{0.78} \std{.09} & \textbf{0.84} \std{.04} & \textbf{0.76} \std{.08} \\
    sequential video  & \textbf{0.998} \std{.000} & \textbf{0.994} \std{.003} & 0.13 \std{.03} & $-0.23$ \std{.02} & 0.08 \std{.02} & $-0.18$ \std{.04} \\
    \bottomrule
  \end{tabular}
\end{table}

\begin{figure}[h]
    \centering
    \includegraphics[width=\linewidth]{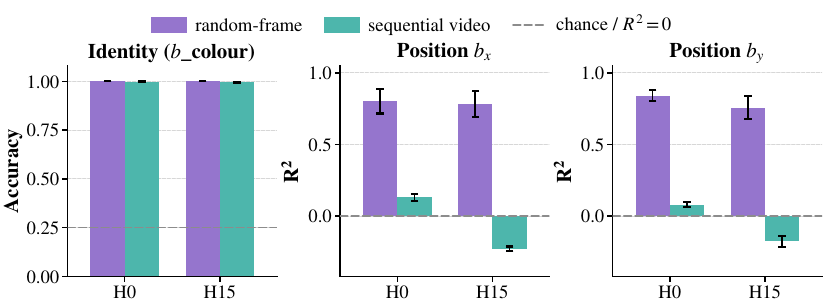}
    \caption{\textbf{\ref{app:passive}: Probe results for passive object $B$} (mean $\pm$ SE, 3 seeds, 212k steps). Object identity reaches ceiling accuracy across both regimes. Position is recovered on randomized backgrounds, but drops to or below $R^2{=}0$ under sequential video.}
    \label{fig:passive_probe}
  \end{figure}

As shown in Table~\ref{tab:passive} and Figure~\ref{fig:passive_probe}, object identity is fully preserved under both backgrounds through 15 rollout steps. However, position recovery depends directly on the background regime: it is accurately decoded under random-frame backgrounds ($R^2 \approx 0.8$), but drops under sequential video ($R^2 \le 0.13$ at H0, dropping below zero after imagination).

This difference stems from the temporal contrastive objective. Because InfoNCE samples negative pairs uniformly across frames, negatives typically come from different episodes. Encoding $B$ lowers negative pair similarity because its attributes vary across episodes while remaining static within each episode. When the background is randomized, $B$ serves as the primary cross-step identifier. On sequential video, however, the temporally coherent background already identifies the episode, diminishing the contrastive utility of encoding $B$'s position.

\begin{figure}[htbp!]
    \centering
    \includegraphics[width=0.85\linewidth]{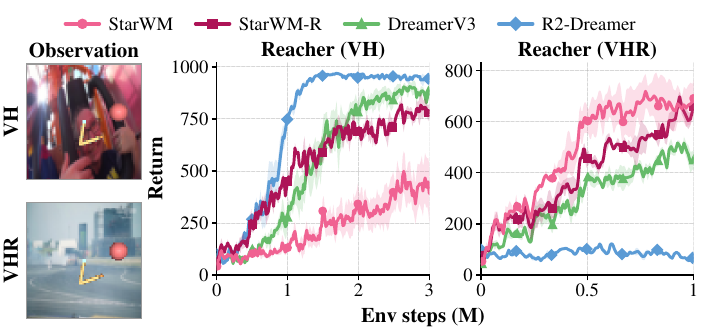}
    \caption{\textbf{\ref{app:passive}: DMC Reacher performance.} \emph{Left:} observation frames showing the two-link arm and static target disc against DAVIS backgrounds (VH: coherent video, VHR: per-step random frames). \emph{Right:} learning curves (mean $\pm$ SE, 3 seeds). On VH, reward augmentation (\starwm{}-R) recovers default \starwm{}'s deficit. On VHR, \starwm{} and \starwm{}-R lead while R2-Dreamer drops.}
    \label{fig:reacher_returns}
\end{figure}

\paragraph{(ii) Fixed target control: DMC Reacher.}
In DMC Reacher, the target is a stationary disc that the agent must reach without directly manipulating it (Figure~\ref{fig:reacher_returns}, left). We evaluate performance across all baselines under both distractor types.

\begin{table}[h]
  \caption{\textbf{\ref{app:passive}: DMC Reacher episode returns} (mean $\pm$ SE, 3 seeds). VH = temporally coherent video background, VHR = per-step randomized frames.}
  \label{tab:reacher}
  \centering
  \begin{tabular}{l|cccc}
    \toprule
    Setting & \starwm{} & \starwm{}-R & DreamerV3 & R2-Dreamer \\
    \midrule
    Reacher VH \,(3M steps)  & 391 \std{136} & 789 \std{33} & 881 \std{34} & \textbf{947} \std{3} \\
    Reacher VHR (1M steps)   & \textbf{690} \std{70} & 657 \std{1} & 450 \std{37} & 66 \std{9} \\
    \bottomrule
  \end{tabular}
\end{table}

Under coherent video (VH), standard \starwm{} scores 391 because the static target lacks both inverse-dynamics gradients and distinct contrastive signals. Adding reward supervision (\starwm{}-R; Appendix~\ref{app:reward_extension}) recovers this gap to 789. Under random backgrounds (VHR), \starwm{} and \starwm{}-R lead at 690 and 657, while R2-Dreamer drops to 66 (Table~\ref{tab:reacher}).

\paragraph{(iii) Conditions for object retention.}
Together with the Ball-in-Cup results (Appendix~\ref{app:reward_extension}), these experiments identify when self-supervised routing retains a task-relevant factor:

\begin{table}[htbp!]
  \caption{\textbf{\ref{app:passive}: Determinants of task-relevant object retention.} Default routing retains features that remain static within an episode and distinct across episodes, provided no competing signal (such as coherent video) already satisfies that criterion.}
  \label{tab:passive_summary}
  \centering
  \small
  \setlength{\tabcolsep}{4pt}
  \begin{tabular}{@{}l|cc|l@{}}
    \toprule
    Task-relevant factor & \makecell{Episode-static\\\& distinctive?} & \makecell{Competing\\signal?} & Default \starwm{} outcome \\
    \midrule
    Passive $B$ position, random bg & \ding{51} & None  & Encoded ($R^2\!\approx\!0.8$) \\
    Passive $B$ position, seq.\ bg  & \ding{51} & Video & Suppressed ($R^2\!\le\!0.13$) \\
    Reacher target, VHR             & \ding{51} & None  & Solved (690) \\
    Reacher target, VH              & \ding{51} & Video & Low return (391); benefits from reward \\
    Ball in flight (Ball-in-Cup)    & \ding{55} & None  & Requires reward routing \\
    \bottomrule
  \end{tabular}
\end{table}

Default routing preserves features that are episode-static and cross-episode distinctive, unless a coherent video background already acts as a dominant episode identifier (Table~\ref{tab:passive_summary}). When self-supervised objectives do not isolate passive elements, reward-conditioned routing incorporates the missing task signals.

\subsection{Delayed Action Effects}
\label{app:delay}

The inverse-dynamics loss in Eq.~\ref{eq:inv} relies on adjacent frames $(t, t{+}1)$, assuming actions take immediate visual effect. When action outcomes are delayed, this adjacent window forces the model to explain a transition before its true cause appears. Here, we show that this sensitivity stems strictly from the supervision window rather than the routing mechanism itself.

\paragraph{The \texttt{ActionDelay}$(k)$ environment.}
We extend NoisyShape such that observations at step $t$ reflect the delayed action $a_{t-k}$ for $k \in \{1, 2, 4\}$, keeping all other task parameters unchanged. We evaluate two supervision windows at each delay $k$:
\textbf{Adjacent} (mismatched) uses $(t, t{+}1)$, targeting an action $a_t$ whose effects are not yet visible.
\textbf{Shifted} (matched) aligns the window to $(t{+}k, t{+}k{+}1)$, capturing the frame pair directly modified by $a_t$ (recovering the standard objective when $k{=}0$).

\paragraph{Evaluation metrics.}
We evaluate two held-out metrics on fresh rollouts: \emph{attention mass on the object} (the fraction and uniform concentration of the top query's attention mask inside the object boundary) and \emph{attribute decodability} from the latent state $[h; z]$ (Appendix~\ref{app:probe_method}). We omit training accuracy of the inverse model, as memorizing static frame--action pairs on the replay buffer easily inflates this metric without reflecting genuine representation quality.

\begin{table}[htbp!]
  \caption{\textbf{\ref{app:delay}: Delayed action effects} at 180k steps (mean $\pm$ SE, 3 seeds). Object coverage is $\approx$6.7\% of the frame; $\times$uniform denotes attention concentration. Color decodability remains near ceiling ($\approx$1.00) across all settings and is omitted. The $k{=}0$ row serves as an un-delayed reference point probed at its final checkpoint ($\approx$145k steps).}
  \label{tab:delay}
  \centering
  \small
  \setlength{\tabcolsep}{4pt}
  \begin{tabular}{@{}ll|cc|cc@{}}
    \toprule
    $k$ & Supervision window & Attn.\ mass & $\times$Uniform & Shape & Size \\
    \midrule
    0 & Reference (no delay) & 0.846 & 11.2$\times$ & 0.999 & 1.000 \\
    \midrule
    1 & Adjacent (mismatched) & 0.439 \std{.033} & 6.6$\times$ & 0.385 \std{.027} & 0.476 \std{.029} \\
    1 & \textbf{Shifted (matched)} & \textbf{0.824} \std{.026} & \textbf{12.3$\times$} & \textbf{0.998} \std{.001} & \textbf{1.000} \std{.000} \\
    2 & Adjacent & 0.352 \std{.093} & 5.3$\times$ & 0.654 \std{.153} & 0.722 \std{.143} \\
    2 & \textbf{Shifted} & \textbf{0.755} \std{.010} & \textbf{11.3$\times$} & \textbf{0.997} \std{.000} & \textbf{0.999} \std{.000} \\
    4 & Adjacent & 0.357 \std{.009} & 6.2$\times$ & 0.448 \std{.039} & 0.504 \std{.028} \\
    4 & \textbf{Shifted} & \textbf{0.725} \std{.020} & \textbf{12.5$\times$} & \textbf{0.994} \std{.001} & \textbf{0.996} \std{.001} \\
    \bottomrule
  \end{tabular}
\end{table}

\paragraph{Results.}
As shown in Table~\ref{tab:delay}, mismatched supervision disperses attention (object mass drops to 0.35--0.44 vs.\ 0.85 without delay), significantly lowering shape and size decodability. Aligning the window to the actual action outcome \textbf{fully restores both metrics}: attention mass recovers to 0.73--0.82 (11--13$\times$ uniform, matching the un-delayed baseline), and probe accuracy for shape and size returns above 0.99 across all delays $k$. 

While mismatched windows induce high variance across runs (e.g., shape decodability at $k{=}2$ is $0.654 \pm 0.153$), their performance consistently lags behind the matched setup. These results confirm that delayed effects do not hinder the routing mechanism itself; aligning the supervision window to the true transition dynamics fully preserves representation learning.
\section{Broader Impacts}
\label{app:broader_impact}

This paper presents foundational research on visual representation learning for model-based reinforcement learning, evaluated entirely on simulated benchmarks (DeepMind Control Suite, DAVIS video backgrounds). The work does not target a specific application or deployed system.

\paragraph{Positive impacts.}
Distraction-robust world models could benefit beneficial applications such as robotic manipulation in cluttered environments, assistive autonomy in real-world settings with varying visual conditions, and sample-efficient learning where collecting clean data is expensive. The architectural separation between deciding \emph{where} to route and \emph{what} to extract may also support interpretability research, since the attention maps provide direct visualization of what the world model is attending to.

\paragraph{Potential negative impacts.}
As with any general-purpose advance in autonomous control, improvements in distraction robustness apply equally to applications with negative dual-use potential (e.g., autonomous weapons systems). We do not see specific harms unique to this work beyond those associated with the broader trajectory of visual RL research.
\newpage
\input{checklist.tex}

\end{document}

%% file: checklist.tex
\section*{NeurIPS Paper Checklist}
\addcontentsline{toc}{section}{NeurIPS Paper Checklist}

\begin{enumerate}

\item {\bf Claims}
    \item[] Question: Do the main claims made in the abstract and introduction accurately reflect the paper's contributions and scope?
    \item[] Answer: \answerYes{}
    \item[] Justification: The abstract and introduction state our main claims and these claims are directly supported by results in §\ref{sec:exp_diagnostic} (probe analysis on NoisyShape and cross-sample image composition) and §\ref{sec:exp_dmcgb} (DMC policy performance).

\item {\bf Limitations}
    \item[] Question: Does the paper discuss the limitations of the work performed by the authors?
    \item[] Answer: \answerYes{}
    \item[] Justification: A dedicated \textbf{Limitations} paragraph in §\ref{sec:conclusion} discusses the limitations of this work.

\item {\bf Theory assumptions and proofs}
    \item[] Question: For each theoretical result, does the paper provide the full set of assumptions and a complete (and correct) proof?
    \item[] Answer:\answerNA{}
    \item[] Justification: The paper presents an architectural and empirical contribution; it does not contain formal theorems or proofs.

    \item {\bf Experimental result reproducibility}
    \item[] Question: Does the paper fully disclose all the information needed to reproduce the main experimental results of the paper to the extent that it affects the main claims and/or conclusions of the paper (regardless of whether the code and data are provided or not)?
    \item[] Answer: \answerYes{}
    \item[] Justification: §\ref{sec:method} describes the architecture (cross-attention module, dual-stream decoder, stop-gradient barriers, training objective) in full. Appendix~\ref{app:algorithm} provides pseudocode (Algorithm~\ref{alg:starwm}) and complete implementation details (encoder interception, multi-head attention, KL annealing schedule, etc.). Appendix~\ref{app:hyperparams} lists all hyperparameters. Appendix~\ref{app:env_design} describes the NoisyShape environment design and DMC video distractor setup. Appendix~\ref{app:baseline} describes baseline implementations.

\item {\bf Open access to data and code}
    \item[] Question: Does the paper provide open access to the data and code, with sufficient instructions to faithfully reproduce the main experimental results, as described in supplemental material?
    \item[] Answer: \answerYes{}
    \item[] Justification: Anonymized code is provided in the supplementary material. Code will also be released on GitHub upon acceptance.

\item {\bf Experimental setting/details}
    \item[] Question: Does the paper specify all the training and test details (e.g., data splits, hyperparameters, how they were chosen, type of optimizer) necessary to understand the results?
    \item[] Answer: \answerYes{}
    \item[] Justification: Appendix~\ref{app:hyperparams} provides complete hyperparameters.  Probe analysis method is described in §\ref{app:probe_method}.

\item {\bf Experiment statistical significance}
    \item[] Question: Does the paper report error bars suitably and correctly defined or other appropriate information about the statistical significance of the experiments?
    \item[] Answer: \answerYes{}
    \item[] Justification: All DMC results report mean and standard error over 3 random seeds (Figure~\ref{fig:dmc_curves}, Table~\ref{tab:dmc_full} in Appendix~\ref{app:full_results}). Probe analysis on NoisyShape also reports mean and standard error over 3 seeds (Table~\ref{tab:noisyshape_full}). Following the reliable RL evaluation framework of~\citet{agarwal2021deep}, we provide \textit{Aggregate metrics with 95\% Stratified Bootstrap CI}, \textit{Probability of Improvement Heatmap} and \textit{Performance Profile} for DMC tasks in Appendix~\ref{app:reliable_rl}


\item {\bf Experiments compute resources}
    \item[] Question: For each experiment, does the paper provide sufficient information on the computer resources (type of compute workers, memory, time of execution) needed to reproduce the experiments?
    \item[] Answer: \answerYes{}
    \item[] Justification: Compute resources are described in Table~\ref{tab:compute_per_alg_mode} in Appendix~\ref{app:baseline}.
    
\item {\bf Code of ethics}
    \item[] Question: Does the research conducted in the paper conform, in every respect, with the NeurIPS Code of Ethics \url{https://neurips.cc/public/EthicsGuidelines}?
    \item[] Answer: \answerYes{}
    \item[] Justification: The research conforms with the NeurIPS Code of Ethics. 

\item {\bf Broader impacts}
    \item[] Question: Does the paper discuss both potential positive societal impacts and negative societal impacts of the work performed?
    \item[] Answer: \answerYes{}
    \item[] Justification: A short broader-impact paragraph appears in Appendix~\ref{app:broader_impact}. The work is foundational research on visual representation learning for model-based RL. 
    
\item {\bf Safeguards}
    \item[] Question: Does the paper describe safeguards that have been put in place for responsible release of data or models that have a high risk for misuse (e.g., pre-trained language models, image generators, or scraped datasets)?
    \item[] Answer: \answerNA{}
    \item[] Justification: The paper does not release pre-trained generative models, scraped datasets, or other assets with foreseeable misuse potential. The released code trains world models on simulated control tasks.

\item {\bf Licenses for existing assets}
    \item[] Question: Are the creators or original owners of assets (e.g., code, data, models), used in the paper, properly credited and are the license and terms of use explicitly mentioned and properly respected?
    \item[] Answer: \answerYes{}
    \item[] Justification: All external assets are properly cited and used in accordance with their licenses: 
    \begin{itemize}
        \item DeepMind Control Suite~\citep{tassa2018deepmind} (Apache 2.0)
        \item DAVIS 2017~\citep{perazzi2016benchmark} (BSD 3-Clause License)
        \item DreamerV3[Original JAX]~\citep{hafner2025mastering} (MIT License); DreamerV3 [PyTorch reimplementation; NM512/dreamerv3-torch on GitHub] (MIT License)
        \item DenoisedMDP~\citep{wang2022denoised} (Creative Commons)
        \item Iso-Dreamer++~\citep{pan2023isodreampp} (MIT License)
        \item R2-Dreamer~\citep{morihira2026rdreamer} (MIT License)
    \end{itemize} . 

\item {\bf New assets}
    \item[] Question: Are new assets introduced in the paper well documented and is the documentation provided alongside the assets?
    \item[] Answer: \answerYes{}
    \item[] Justification: We introduce the NoisyShape diagnostic environment (described in Appendix~\ref{app:shapeenv}) and the \starwm{} implementation. Both will be released with upon acceptance; anonymized versions are provided in supplementary material with documentation.

\item {\bf Crowdsourcing and research with human subjects}
    \item[] Question: For crowdsourcing experiments and research with human subjects, does the paper include the full text of instructions given to participants and screenshots, if applicable, as well as details about compensation (if any)? 
    \item[] Answer: \answerNA{}
    \item[] Justification: The research does not involve crowdsourcing or human subjects.

\item {\bf Institutional review board (IRB) approvals or equivalent for research with human subjects}
    \item[] Question: Does the paper describe potential risks incurred by study participants, whether such risks were disclosed to the subjects, and whether Institutional Review Board (IRB) approvals (or an equivalent approval/review based on the requirements of your country or institution) were obtained?
    \item[] Answer: \answerNA{}
    \item[] Justification: The research does not involve crowdsourcing or human subjects.

\item {\bf Declaration of LLM usage}
    \item[] Question: Does the paper describe the usage of LLMs if it is an important, original, or non-standard component of the core methods in this research? Note that if the LLM is used only for writing, editing, or formatting purposes and does \emph{not} impact the core methodology, scientific rigor, or originality of the research, declaration is not required.
    \item[] Answer: \answerNA{}
    \item[] Justification: LLMs were not used as a component of the core methods (training, evaluation, or analysis). LLMs were used only for writing assistance (editing, formatting) and coding assistance.

\end{enumerate}